\documentclass[10pt,twocolumn,letterpaper]{article}

\usepackage{iccv}
\usepackage{times}
\usepackage{epsfig}
\usepackage{graphicx}
\usepackage{amsmath}
\usepackage{amssymb}
\usepackage{booktabs}
\usepackage{bm}
\usepackage{multirow}
\usepackage{xurl}
\usepackage{makecell}

\usepackage[toc,page]{appendix}

\DeclareUnicodeCharacter{2212}{-}

\usepackage{hyperref}
\hypersetup{
    colorlinks=true,            
    linkcolor=blue,             
    filecolor=magenta,          
    urlcolor=cyan,              
    pdftitle={Overleaf Example},
    pdfpagemode=FullScreen,
    }

\usepackage{authblk}

\iccvfinalcopy 

\def\iccvPaperID{****} 

\ificcvfinal\fi

\begin{document}

\title{Unsupervised Evaluation of Out-of-distribution Detection: \\A Data-centric Perspective}

\author[1, 2]{Yuhang Zhang}
\author[1]{Weihong Deng}
\author[2]{Liang Zheng}

\affil[1]{Beijing University of Posts and Telecommunications}
\affil[2]{Australian National University}
\affil[ ]{\textit {\{zyhzyh, whdeng\}@bupt.edu.cn, liang.zheng@anu.edu.au}}

\maketitle
\ificcvfinal\thispagestyle{empty}\fi

\begin{abstract}
   Out-of-distribution (OOD) detection methods assume that they have test ground truths, i.e., whether individual test samples are in-distribution (IND) or OOD. However, in the real world, we do not always have such ground truths, and thus \textbf{do not} know which sample is correctly detected and cannot compute the metric like AUROC to evaluate the performance of different OOD detection methods. In this paper, we are the first to introduce the unsupervised evaluation problem in OOD detection, which aims to evaluate OOD detection methods in real-world changing environments without OOD labels. 
   We propose three methods to compute Gscore as an unsupervised indicator of OOD detection performance. We further introduce a new benchmark Gbench, which has 200 real-world OOD datasets of various label spaces to train and evaluate our method. Through experiments, we find a strong \textbf{quantitative} correlation between Gscore and the OOD detection performance.
   Extensive experiments demonstrate that our Gscore achieves state-of-the-art performance. Gscore also generalizes well with different IND/OOD datasets, OOD detection methods, backbones and dataset sizes. We further provide interesting analyses of the effects of backbones and IND/OOD datasets on OOD detection performance. The data and code will be available.

\end{abstract}

\begin{figure}[t]
  \centering
  \includegraphics[width=1.\linewidth]{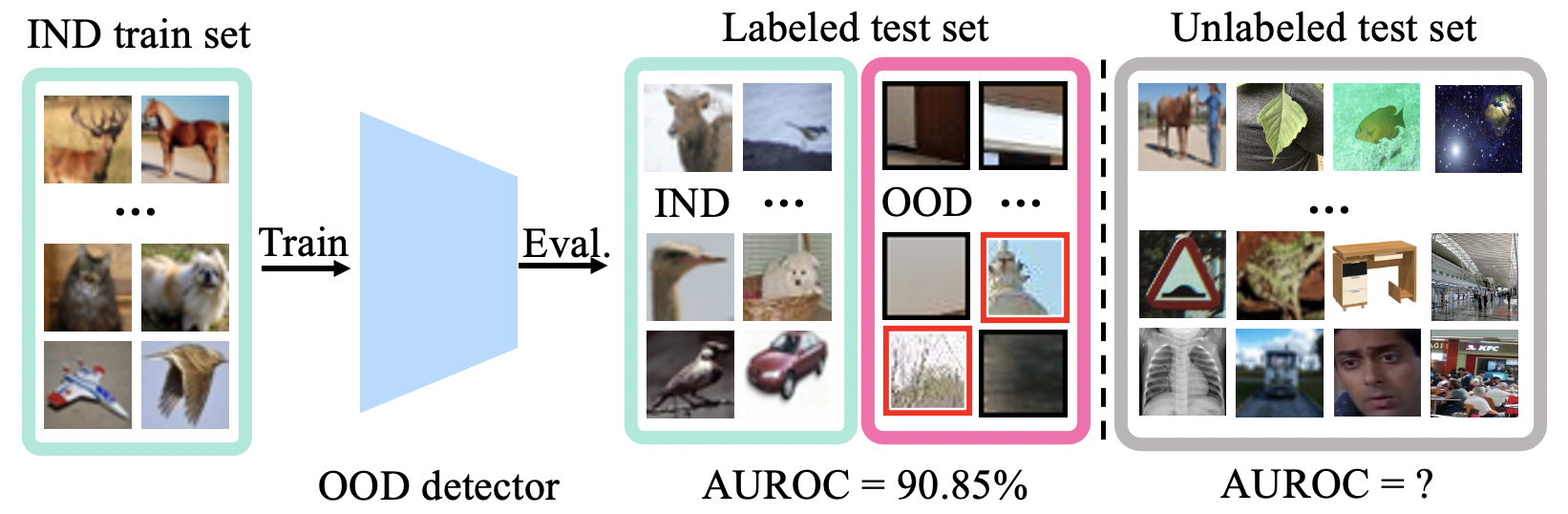}
    \caption{An illustration of unsupervised evaluation of OOD detection. Normally we evaluate OOD detection methods on labeled test sets, so we know which sample is correctly detected and can compute OOD detection performance like AUROC. However, when we evaluate OOD detection models in the real world, we do not have OOD labels and thus do not know which sample is correctly detected. Under this practical scenario, AUROC can no longer be computed. In this paper, we aim to predict the performance of different OOD detection methods \emph{without} OOD labels.}
    \label{auroce}
\end{figure}

\section{Introduction}
\label{sec:intro}
Out-of-distribution (OOD) detection aims to detect image objects belonging to a different label space from the training categories, which is vital for the safe deployment of computer vision systems. For this problem, extensive efforts are made to find discriminative OOD scores~\cite{hendrycks2017a, liang2018enhancing, NEURIPS2020_f5496252}, tune with OOD validation sets~\cite{hendrycks2019oe, yu2019unsupervised, yang2021semantically} or regularize training of OOD detectors~\cite{devries2018learning, hsu2020generalized, tack2020csi, huang2021mos, du2022vos}, \textit{etc}. 

Existing works in this community typically perform evaluation on a test set \textit{with ground truths}, which means they already know a sample belongs to in-distribution (IND) or OOD. Specifically, an OOD confidence score is usually computed for each test sample, and thresholds are used to make IND or OOD predictions. By comparing the predictions with ground truths, OOD detection performance (\textit{e.g.}, FPR@TPR95~\cite{hendrycks2017a}, AUROC~\cite{liang2018enhancing}) can be computed.

However, in deployment, the above evaluation routine faces challenges. A major one is that we would not have test ground truths, and even if we manage to have them for one environment it is prohibitively costly to annotate a test set every time we meet a new test environment. Therefore, we can no longer evaluate our system as we normally do. Besides, technically speaking, the system would encounter test samples from a wide range of OOD categories, instead of those commonly used label spaces like CIFAR-100 \cite{krizhevsky2009learning}, SVHN \cite{Netzer2011ReadingDI}. This emphasizes the model evaluation in real-world environments: some categories are easier to be detected as outliers, while others more difficult; so detection performance on certain test sets like CIFAR-100 or SVHN does not reflect the detection performance on others.

Addressing these problems brings important benefits. From the perspective of real-world deployment, it allows us to predict system failure in different environments, without having to label test samples. From a scientific perspective, we will be able to better understand the properties of data from various label spaces and answer questions like what OOD datasets are more difficult to detect for a given detector and IND dataset. 

\begin{figure}[t]
  \centering
  \includegraphics[width=1.\linewidth]{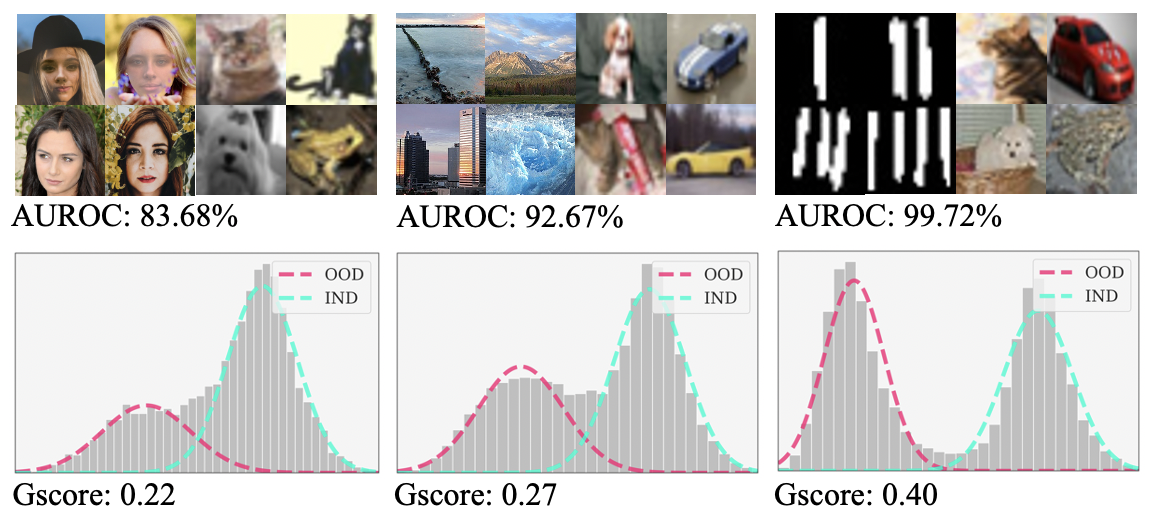}
  \caption{Our motivation for Gscore design. In the upper part, we show three test sets, each composed of OOD (left) and IND (right) samples. Using the ODIN \cite{liang2018enhancing} detection method, we observe for each test set a bimodal distribution of the OOD scores. From left to right, we find that OOD detection performance (AUROC) increases when the two distribution components are better separated. Therefore, we design Gscore to reflect such separability, the value of which increases from left to right. We find a strong \emph{quantitative} correlation between our proposed Gscore and OOD detection performance through extensive experiments, which enables performance prediction without OOD labels.}
\label{motivation}
\end{figure}

In light of the above discussions, we are the first to evaluate OOD detection methods on \textit{unlabelled test sets of various label spaces}, which is illustrated in Fig. \ref{auroce}. In designing such an unsupervised evaluation method, we are mainly motivated by the observation that the OOD confidence score distribution on the test set is usually bimodal Fig.~\ref{motivation}. Intuitively, a high separability between the two components (modalities) indicates that IND and OOD data are well separated, meaning high OOD detection performance. To reflect such separability, we propose Gscore (`G' for `generalization') to represent the distribution difference between IND and OOD test data. We further propose three methods Kmeans, GMM and Unilateral Density Estimation (UDE) to compute the Gscore. 

Though the motivation seems intuitive, we claim that previous works only provide a general feeling that large distribution difference leads to better performance. Our contribution lies in that we are the first to \emph{quantitatively} connect an unsupervised proxy with OOD detection performance, which enables performance prediction without OOD labels. We utilize a regression model to fit the correlation between Gscore and OOD performance. When facing unlabelled test set, we only need to compute Gscore, then the trained regression model will predict the OOD performance. The other main contribution is that we propose a new benchmark named Gbench, which contains \emph{200} real-world OOD datasets of various label spaces. We train and evaluate our proposed method in the Gbench and achieves state-of-the-art performance compared with other methods.

Our main observation is that we find a strong correlation between Gscore and OOD detection performance, which enables the regression model to fit the relationship between them. We further validate that this correlation still exists when we use different OOD detection methods, IND/OOD datasets, and test set sizes, which strongly supports unsupervised performance evaluation of different methods on various IND/OOD data without labels. Moreover, Gbench reveals some interesting observations about the effect of different backbones, IND/OOD datasets. For example, the classification performance shows a positive correlation with the OOD detection performance. In another example, if IND data changes, OOD detection performance on different OOD data changes drastically. Our contributions are summarized below. 

\begin{itemize}
\item We are the first to propose the problem of unsupervised evaluation of OOD detection. We propose three methods to compute Gscore to solve this problem and we find a strong quantitative correlation between the proposed Gscore and OOD detection performance, which enables OOD detection performance prediction even without OOD labels.
\item We introduce Gbench, a suite of 200 OOD datasets of various label spaces. It allows us to validate the quantitative relationship between Gscore and OOD detection performance and evaluate the performance of different unsupervised evaluation methods. Gbench can also be utilized as a benchmark to test the generalization ability of existing OOD detection methods.
\item Extensive experiments validate that our proposed Gscore achieves state-of-the-art performance under different OOD detetion methods, IND/OOD datasets and test set sizes. We further provide interesting findings utilizing Gbench, like the effect of different IND/OOD datasets or backbones on the OOD detection performance.
\end{itemize}

\section{Related Work}
\label{related work1}

\textbf{Out-of-distribution (OOD) detection} 
has been extensively studied~\cite{hendrycks2017a, liang2018enhancing, NEURIPS2020_f5496252, lee2018simple, sastry2020detecting, huang2021importance, sun2021react, hendrycks2019scaling, wang2022vim, sun2022knnood, hendrycks2019oe, devries2018learning, hsu2020generalized, tack2020csi, huang2021mos, du2022vos, hendrycks2019oe, yu2019unsupervised, yang2021semantically, yang2022openood, sastry2020detecting}. MSP~\cite{hendrycks2017a} is a widely used baseline, which directly uses the maximum softmax probability to detect OOD samples. ODIN~\cite{liang2018enhancing} utilizes temperature scaling and adversarial perturbations to widen the gap between IND samples and OOD samples, which makes OOD detection easier. ENERGY~\cite{NEURIPS2020_f5496252} uses energy scores that are theoretically aligned with the probability density of the inputs, which are less susceptible to the overconfidence problem of softmax scores. Some other methods employ OOD sets for training~\cite{hendrycks2019oe, yu2019unsupervised, yang2021semantically} or train-time regularization~\cite{devries2018learning, hsu2020generalized, tack2020csi, huang2021mos, du2022vos}.  
However, these methods all evaluate their systems on test sets with ground truths. \emph{In this paper, we study the unsupervised evaluation problem, providing an orthogonal perspective to the OOD detection field.}

\textbf{Unsupervised model evaluation} aims to predict model accuracy when %
we cannot acquire labels~\cite{deng2021labels, chen2021detecting, deng2021does, garg2022leveraging, sun2021ranking, joyce2021framework, risser2022can, li2022estimating, sun2021label}. Deng and Zheng~\cite{deng2021labels} utilize the Frechet distance between training and test sets, which can be used as a proxy for classification accuracy. 
Guillory \textit{et al.} \cite{guillory2021predicting} propose difference of confidences, an uncertainty-based indicator to classification accuracy. Saurabh \textit{et al.} \cite{Saurabh2022} propose to learn a confidence threshold and use the proportion of unlabelled examples exceeding the threshold as a proxy for model accuracy. Ji \textit{et al.}~\cite{Ji2020} first calibrate the model which gives better uncertainty scores for accuracy estimation. \emph{The above methods are all in the generic image classification field, 
while in this paper, we  are the first to propose an indicator that is specifically designed for OOD detection.}

\section{Preliminary and Problem Definition}
\subsection{OOD detection revisit}
\label{sec:revisit}
Out-of-distribution (OOD) detection methods predict whether a test sample is from a different distribution from the training set. An OOD detection model is trained on a labeled in-distribution (IND) dataset $D^{train}_{ind} = \{(\bm{x}_i, y_i)\}^{M}_{i=1}$, where $\bm{x}_i$ is an image, $y_i$ is its \emph{class label}, and $M$ is the number of images. 
In the testing phase, IND test set $D^{test}_{ind} = \{\bm{x}_j\}^{N}_{j=1}$ and OOD test set $D^{test}_{ood} = \{\tilde{\bm{x}}_j\}^{N}_{j=1}$ are 
mixed to form $D^{test} = \{(\bm{x}_k, y_k)\}^{2N}_{k=1}$, where $\bm{x}_k$ is an image, $y_k$ is the \emph{OOD label} indicating whether image $\bm{x}_k$ is from $D^{test}_{ind}$ or $D^{test}_{ood}$. Usually, an OOD score which represents the probability of a sample belonging to the IND is acquired, denoted as $S = \{s_k\}^{2N}_{k=1}$. After selecting a score threshold $t$, $D^{test}$ can be divided into two parts $D^{test}_{high}$ and $D^{test}_{low}$. $D^{test}_{high}$ is the predicted IND dataset with $s_k > t$ and vice versa. They are compared to $D^{test}_{ind}$ and $D^{test}_{ood}$ to evaluate the performance of OOD detection methods. The most widely used metrics are FPR@TPR95 and AUROC. 1) FPR@TPR95 measures the false positive rate (FPR) when the true positive rate (TPR) is 95\%. Lower scores indicate better performance. 2) AUROC is the area under the Receiver Operating Characteristic (ROC) curve, which represents the probability that an IND sample has a higher OOD score than an OOD sample. Higher is better.

\subsection{Problem definition - unsupervised evaluation} 
We focus on the evaluation procedure, so the training phase has the same notations and process described in Section \ref{sec:revisit}. In evaluation, instead of having a test set with ground truth labels $D^{test} = \{(\bm{x}_k, y_k)\}^{2N}_{k=1}$, we assume an unlabelled test set $D^{test} = \{\bm{x}_k\}^{2N}_{k=1}$ is given. We aim to predict OOD detection performance $p$, such as AUROC, of any given OOD detection method on different $D^{test}$. Traditional OOD detection methods fail to predict $p$ as we do not have OOD labels and thus do not know which test sample is correctly detected and which one is not. In this paper, we find an unsupervised performance indicator (Gscore) to solve this problem. We first calculate Gscore $s$ without OOD labels as $s = f(D^{test})$, where $f$ is the function to generate Gscore. After that, we utilize a trained regression model $g$ to predict the OOD detection performance $p$ only given Gscore $s$, following $p = g(s)$. The regression model $g$ is trained on our constructed meta-train sets $D^{meta}$ to regress the relationship between Gscore $s$ and detection performance $p$. Note that meta-train sets have no overlap with the unlabelled test sets, details are illustrated in section \ref{details}.

\section{Gbench: A Benchmark of 200 Datasets}
As we need to quantitatively connect Gscore $s$ with detection performance $p$, we construct many different meta-train sets $D^{meta}$. We collect 200 datasets as a dataset suite named Gbench. We utilize 150 of them as meta-train sets and the other as test sets. Note that the meta-train sets have no overlap with the test sets. All of these datasets are publicly available, either commonly used in computer vision research or released in Kaggle competitions. Some of the datasets, such as LSUN and Tiny-ImageNet are very commonly used in the OOD detection community. Other datasets are much less studied but have interesting content, such as medical images, crack detection, butterflies, faces, and satellite images. A complete list of the datasets is provided in the supplementary material. 

Moreover, Gbench has high diversity, in the sense that OOD detection performance on its 200 datasets varies significantly. \emph{Gbench contributes to the OOD detection community as it can also be utilized to evaluate the generalization ability of existing OOD detection methods.} In the supplementary material, we present the AUROC distribution on 150 datasets\footnote{ A split of Gbench used for regression training}. The ODIN method is used for detection. We observe that the AUROC ranges from around 57\% to 100\% \footnote{ Random guess yields an AUROC of 50\%.}, indicating that datasets in Gbench have a wide span in their OOD difficulty.

\section{Proposed Approach}
\subsection{Gscore: a measurement of the separability of IND and OOD test data}
\label{Gscore1}
\textbf{Motivation of the design of Gscore.} We are mainly inspired by Fig.~\ref{motivation}, which shows that well separated IND and OOD data indicate higher OOD detection performance. To \emph{quantitatively} reflect such separability, we design a proxy named Gscore. 
Specifically, we observe that the OOD scores on the test data usually form a bimodal distribution, and a few examples are shown in Fig.~\ref{motivation}. If we could model the distributions of IND and OOD data, separately, we will be able to compute Gscore by measuring the distribution difference under certain metrics. Finally, we connect Gscore with OOD detection performance through a regression model. Thus, when facing unlabelled data, we can compute Gscore to estimate the corresponding detection performance. We describe distribution modeling and distribution difference measurement below. 

\textbf{Distribution modeling of IND and OOD test data.} 
We can use unsupervised methods like Kmeans \cite{krishna1999genetic} and Gaussian mixture model (GMM) \cite{permuter2006study} to model the distribution of OOD scores on test sets. This modeling process can be further simplified because there are only two types of test data, \textit{i.e.}, IND and OOD. Therefore, we set the number of components in Kmeans and GMM to 2. 

\begin{itemize}
    \item Kmeans starts with two random centroids and iteratively updates the centroids $\mu_{ind}$ and $\mu_{ood}$.
    \item GMM gives us means $\mu_{ind}$ and $\mu_{ood}$ and variances $\sigma_{ind}$ and $\sigma_{ood}$ of IND and OOD data, respectively.
\end{itemize}

Apart from Kmeans and GMM, we propose a Unilateral Density Estimation (UDE) method. Specifically, as all OOD detection methods \cite{hendrycks2017a, liang2018enhancing, NEURIPS2020_f5496252, lee2018simple,  huang2021importance, sun2021react, hendrycks2019scaling, wang2022vim, sun2022knnood, hendrycks2019oe, devries2018learning,  tack2020csi, huang2021mos, du2022vos} train their models on in-distribution (IND) dataset, we set up a little part of the train set as validation set to estimate the confidence score distribution of IND data as a single Gaussian $p$ , denoted as 
$p = \frac{1}{\sigma^{val}\sqrt{2\pi}}\exp(-\frac{(x-\mu^{val})^2}{2 {\sigma^{val}}^2})$. Experiments in the supplementary material show that the size of validation set has little effect on the performance of our method. Given an unlabelled test set, we measure the probability of each sample generated by $p$ following $s_p = \exp(-\frac{(x-\mu^{val})^2}{2 {\sigma^{val}}^2})$,
$s_p \in [0,1]$, and use a threshold to divide all the test samples into IND and OOD subsets according to the score $s_p$. The threshold choice is illustrated in Section~\ref{regressionto}. Finally, we use two single Gaussian distributions to separately model the OOD scores of IND and OOD subsets and obtain the corresponding distribution parameters $\mu_{ind}$, $\sigma_{ind}$ and $\mu_{ood}$, $\sigma_{ood}$. 

\textbf{Measuring distribution difference between IND and OOD test data.} 
\label{distances}
After obtaining the distribution parameters, we compute the distribution difference between IND and OOD test data, using existing metrics. The resulting distance is named Gscore. Specifically, we could use $\ell_2$ distance, KL divergence or Wasserstein distance. For Kmeans, it outputs the 2 centroids, denote as $\mu_{ind}$ and $\mu_{ood}$. L2 distance is computed as $\ell_2 = |\mu_{1} - \mu_{2}|$, where $\mu_{1}$, $\mu_{2}$ are $\mu_{ind}$, $\mu_{ood}$. GMM and UDE return the mean and standard variance of the two distributions, denoted as $\mu_{ind}$, $\sigma_{ind}$, $\mu_{ood}$, $\sigma_{ood}$. KL divergence distance is computed using: 
\begin{equation}
KL = \log \frac{\sigma_{1}}{\sigma_{2}} + \frac{\sigma_{2}^2+(\mu_{1}-\mu_{2})^2}{2\sigma_{1}^2}-\frac{1}{2}.
\end{equation}
Wasserstein distance is computed by:
\begin{equation}
W = ||\mu_{1}-\mu_{2}||^2+||\sigma_{1}-\sigma_{2}||^2,
\end{equation}
where $\mu_{1}$, $\mu_{2}$ can take the value of $\mu_{ind}$, $\mu_{ood}$, respectively and $\sigma_{1}$, $\sigma_{2}$ can be $\sigma_{ind}$, $\sigma_{ood}$, respectively. Because KL divergence is asymmetric, $\sigma_{1}$, $\sigma_{2}$ can also be $\sigma_{ood}$, $\sigma_{ind}$.

\subsection{Regression to quantitatively connect Gscore and OOD detection performance}
\label{regressionto}
\textbf{Training.}
We aim to train a regression model that uses Gscore as input and the latent truth OOD detection performance as target. When facing with unlabelled test sets, the pretrained regression model will predict the OOD detection performance only given the unsupervised indicator Gscore. 

Formally, assume we have $N$ meta-train sets. Each meta-train set $D_i$ can be denoted as $\{D^{ind}, D^{ood}_i\}$, where $D^{ind}$ can be the CIFAR-10 training set, and $D^{ood}_i$ is the $i$th OOD dataset. 
We first extract Gscore $Gs_i$ from the meta-train set, and then get the corresponding latent truth OOD detection performance $p_i$ for the training of the regression model.
The linear regression model $f$ is written as, 
\begin{equation}
    p_i = f(Gs_i | \theta_1, \theta_0) = \theta_1 Gs_i + \theta_0,
\end{equation}
where $\theta_1$ and $\theta_0$ are regression parameters. We use a standard least square loss to optimize this regression model.

\textbf{Validation.} Our system has only one hyperparameter, which is the threshold $\tau$ when we use UDE for distribution modeling. We select its value on the meta-train sets $D_i, i=1,...,N$, by enumerating its possible values from 0 to 1 with an interval of 0.1. After acquiring the initial $\tau$, we can repeat the process from $\tau-0.5$ to $\tau+0.5$ to choose a finer $\tau$ with an interval of 0.01. More specifically, every time we train a regression model $f$ with a certain value of $\tau$, we record the final loss value. We decide the optimal value $\tau_{op}$ such that the regression model $f$ has the minimal train loss value. We evaluate our model using $\tau_{op}$. Note that GMM is also with a validation process while Kmeans is not. More details are in the supplementary material. 

\textbf{Testing.}
Given an unlabelled test set $D = \{\bm{x}\}$, we first compute its Gscore $Gs$. We then predict the detector performance $p$ on $D$ by $p = f(Gs)$. Note that our meta-train sets are composed by part of CIFAR-10 \emph{train set} and OOD datasets, while the unlabelled test set is composed by CIFAR-10 \emph{test set} and different OOD datasets from meta-train sets, \emph{none of them are seen during the training phase}.

\begin{table*}[t]
\centering
\caption{The performance of different unsupervised evaluation methods on 50 real-world unlabelled test sets. We test on four OOD detection methods (MSP, ODIN, ENERGY, MLS) and two OOD detection metrics (FPR@TPR95, AUROC). We report the deviation between the predicted OOD detection metric and the latent truth, measured by RMSE (\%), \emph{the lower the better}. We use different random seeds and each experiment is carried out 5 times, and the mean and standard deviation of RMSE are reported. \emph{The best results are shown in bold, and our proposed Gscore achieves the best performance under all settings.}}
\label{autoeval}
\resizebox{\linewidth}{!}{

\begin{tabular}{ll|cccc|cccc}
\Xhline{1.2pt}
 \multicolumn{2}{l|}{\multirow{2}{*}{Unsup. Eval. Methods} }            & \multicolumn{4}{c|}{FPR@TPR95}                 & \multicolumn{4}{c}{AUROC}                     \\ \cline{3-10} 
 &   & MSP       & ODIN       & ENERGY    & MLS       & MSP       & ODIN      & ENERGY    & MLS       \\ \hline
 
\multicolumn{2}{l|}{Confidence score (Baseline)} & 71.86±0.56 & 40.33±2.13  & 46.33±1.87 & 47.30±1.53 & 17.95±0.54 & 12.75±1.20 & 13.67±0.86 & 13.65±0.84 \\ 
\multicolumn{2}{l|}{Auto evaluation} & 12.69±0.61& 12.69±0.61  & 12.69±0.61 & 12.69±0.61 & 5.92±0.33 & 5.92±0.33 & 5.92±0.33 & 5.92±0.33 \\  \hline
\multicolumn{2}{l|}{Gscore (Kmeans + l2)}       & 8.73±1.22 & 10.97±0.86 & 6.79±0.73 & 6.93±0.77 & 5.32±0.75 & 5.25±0.38 & 5.01±0.75  & 5.17±0.79 \\
\multicolumn{2}{l|}{Gscore (GMM + Wasserstein)} & 4.41±0.66 & 6.89±0.46  & 7.28±0.79 & 6.30±0.68 & 4.81±0.55 & 4.53±0.41 & 5.44±0.84 & 4.85±0.54 \\ 
\multicolumn{2}{l|}{Gscore (UDE + Wasserstein)} & \textbf{3.46±0.63} & \textbf{4.50±0.23}  & \textbf{4.39±0.55} & \textbf{4.52±0.57} & \textbf{3.64±0.27} & {\textbf{3.86±0.32}} & \textbf{4.02±0.51} & \textbf{4.08±0.52} \\ \Xhline{1.2pt}
\end{tabular}}
\vspace{-2mm}
\end{table*}

\subsection{Discussion}
\textbf{Difference between existing methods} Traditional OOD detection methods cannot evaluate their performance without OOD labels, while our proposed method solves this problem. The most related work to our method is the unsupervised evaluation of image classification task \cite{deng2021labels}, while it is not suitable for OOD detection task. It computes the feature distance between unlabelled test dataset and the original dataset to represent the hardness of the test dataset. However, we claim that feature distance do not reflect the performance of different OOD detection methods. Furthermore, \cite{deng2021labels} can only deal with datasets with the same label space while our proposed method is agnostic to the label spaces of different OOD datasets. We quantitatively compare with \cite{deng2021labels} in section \ref{firstresults}.

\textbf{Kmeans / GMM \textit{vs.} UDE.} Kmeans and GMM are easy to use, and the number of components is predefined as 2 for both methods. However, their performance degrades when IND and OOD test data overlap much or the IND:OOD sample numbers are imbalanced (shown in the supplementary material). In comparison, UDE achieves better performance under both conditions.

\textbf{Model-centric, unsupervised evaluation.} This paper mainly studies the scenarios where the test set undergoes changes, which is a data-centric problem. It would be interesting to predict the performance of numerous OOD detectors on a fixed test set, rendering a model-centric perspective. We emphasize that these two problems rely on the characteristics of data and model, respectively, and are thus different in nature. We would like to study the model-centric problem in future work.

\textbf{Unsupervised evaluation in batches instead of in test sets.} In deployment, it would be easier to obtain a batch of test samples than an entire test set. As to be shown in section \ref{analysis}, our method is still effective on small test sets of  50 samples, which is equivalent to mini-batches.

\section{Experiments and Analysis}
\label{experiments}

\subsection{Implementation details}
\label{details}
By default, CIFAR-10 \cite{krizhevsky2009learning} is treated as IND data. We split CIFAR-10 train set into 5 partitions, 4 for training the OOD detector backbone, and the rest as the meta-train IND set. We use the training splits to train a strong backbone model DLA~\cite{yu2018deep} and 4 different OOD detectors, including MSP~\cite{hendrycks2017a}, ODIN~\cite{liang2018enhancing}, ENERGY~\cite{NEURIPS2020_f5496252} and MLS~\cite{hendrycks2019scaling}. 
The classification accuracy of the DLA \cite{yu2018deep} model on the CIFAR-10 test set is 93\%. 

We train a regression model on meta-train sets to solve the problem of unsupervised evaluation of OOD detection. We randomly choose 150 OOD datasets from Gbench and mix them with the meta-train IND set to construct 150 meta-train sets. For evaluation of the trained regression model, we use the rest 50 OOD datasets of Gbench and mix with the CIFAR-10 test set to form 50 test sets. \emph{Notice that the test sets have no overlap with the meta-train sets.} We compare the predicted OOD detection performance with the ground truth performance on the 50 test sets and compute the root mean squared error (RMSE) as the evaluation metric. All experiments are conducted on a server with AMD 2950X CPU and 4 NVIDIA RTX 2080Ti GPUs.

If other IND datasets such as Clothing1M \cite{xiao2015learning} are used, we make sure that the datasets that have class overlap with the IND dataset are manually removed from the datasets of Gbench, details are in the supplementary material.

\begin{figure*}[t]
  \centering
  \includegraphics[width=1.\linewidth]{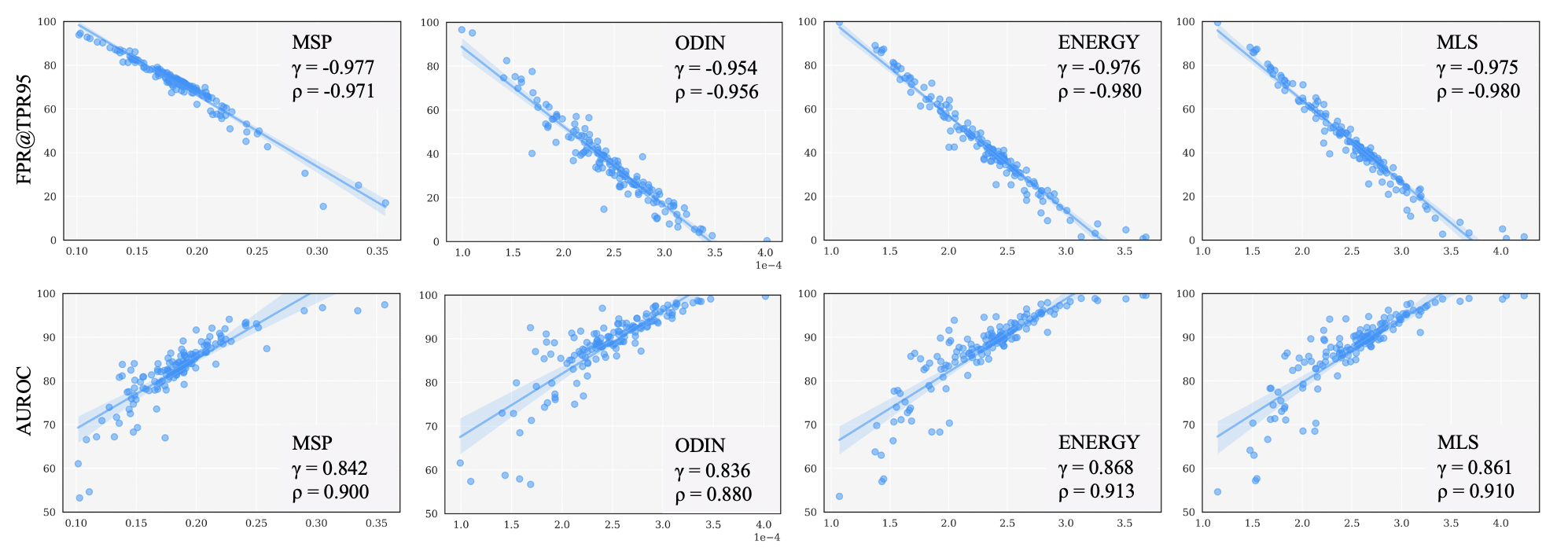}
    \caption{Strong correlation between Gscore and OOD detection performance. We plot results on 150 OOD datasets from a training split of Gbench and study four detection methods (MSP, ODIN, ENERGY, MLS) and two detection performance metrics FPR@TPR95 (\%) and AUROC (\%). $\gamma$ and $\rho$ represent Pearson Correlation Coefficient and Spearman Correlation Coefficient, respectively. The strong correlation allows us to use Gscore to estimate OOD detection performance on various unlabelled test sets.}
    \vspace{-2mm}
    \label{linear relationship}
\end{figure*}

\begin{figure*}[t]
  \centering
  \includegraphics[width=1.\linewidth]{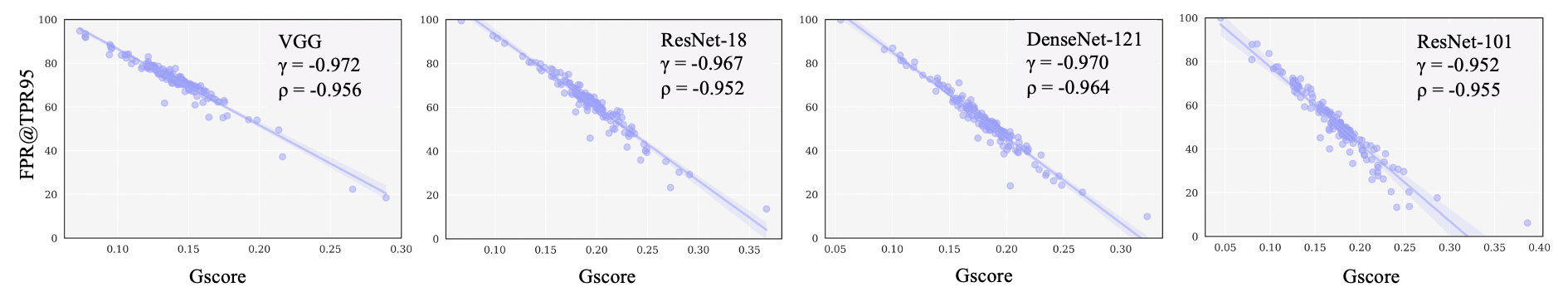}
    \caption{Strong correlation between Gscore and FPR@TPR95 (\%) performance holds across different backbones. We use VGG, ResNet-18, DenseNet-121 and ResNet-101 as the backbeone in the figures. We use the MSP method.}
    \vspace{-4mm}
    \label{different backbones}
\end{figure*}

\subsection{Main results}
\label{firstresults}
\textbf{Our algorithm achieves state-of-the-art performance under various test sets.} We evaluate our method on 50 randomly selected OOD test sets and 4 different OOD detectors. Each experiment is repeated for 5 times with different train/test splits. We report the mean and standard deviation of RMSE on the 50 OOD test sets in Table~\ref{autoeval}. Our baseline utilizes confidence score to estimate the OOD labels, if confidence score is over 0.9, we label the sample as IND. We also compare with the state-of-the-art auto evaluation method in image classification~\cite{deng2021labels}. We have three major observations. 

\textbf{First}, all our three methods outperform the baseline and auto evaluation~\cite{deng2021labels}, our UDE outperforms them by large margins. Though confidence score (baseline) performs well in image classification field, it fails in unsupervised evaluation of OOD detection, as confidence score cannot reflect the performance of OOD detectors. Auto evaluation~\cite{deng2021labels} utilizes Frechet distance to measure the distance of \textbf{features} to indicate the hardness of classification task. While in Table~\ref{autoeval} we have shown that measuring feature distance is not suitable for OOD detection tasks, as the feature distance is not strongly related to the OOD detection performance like OOD scores. Furthermore, auto evaluation considers one dataset as a distribution while it is not suitable for OOD detection as one OOD test set contains two distributions.

\textbf{Second}, Gscore obtained by UDE and Wasserstein distance achieves the best result under all conditions. For example, the RMSE scores in predicting FPR@TPR95 and AUROC are 3.46±0.63\% and 3.64±0.27\%, respectively, under the MSP detector, which means in real-world OOD test sets, our method can predict the OOD detection performance with around 3.5\% error even without OOD labels. 

\textbf{Third}, for different OOD detectors, our performance prediction results are consistently good. For example, when predicting AUROC, the RMSE scores for the four different detectors are 3.64±0.27\%, 3.86±0.32\%, 4.02±0.51\%, and 4.08±0.52\%, respectively, which are rather stable. 

We also compare with a few variants, characterized by Kmeans / GMM with other distance metrics, each with its best distance metric. Results using other distance metrics in Section \ref{distances} are shown in the supplementary material. We observe that UDE+Wasserstein is the best. The reason lies in that UDE leverages a part of meta-train IND set as validation set, and thus can better estimate the two distributions.

\textbf{The strong correlation between Gscore and OOD detection performance}. We now verify that the good estimates are due to the effectiveness of Gscore. We use a random Gbench split and plot the OOD detection performance (FPR@TPR95 and AURUC) on the 150 meta-train sets against their Gscores. Four detection methods are shown, and IND dataset is CIFAR-10. Results are presented in Fig.~\ref{linear relationship}. Pearson Correlation Coefficient $\gamma$~\cite{benesty2009pearson} and Spearman Correlation Coefficient $\rho$~\cite{kendall1948rank} are used to measure correlation linearity and monotonicity.  
Both range from [$-1$, $1$] and a value closer to $−1$ or $1$ indicates a strong negative or positive correlation, respectively. 

We clearly observe that Gscore has a strong correlation with FPR@TPR95 and AUROC. Across all the eight figures, the mean absolute values of $\gamma$ and $\rho$ are 0.911 and 0.936, respectively. Such strong correlation under different OOD detection methods is very close to the linear relationship, which supports evaluation on unlabelled test sets. 

We also find Gscore generally has a stronger correlation with FPR@TPR95 than AUROC, with the mean absolute value of $\gamma$ as $0.971$ versus $0.852$ ($\rho$ has a similar trend). The reason might be that FPR@TPR95 measures FPR at a certain OOD score threshold, while AUROC considers all the possible thresholds. Thus, AUROC represents the entire shape of the two distributions, while FPR@TPR95 mainly represents the degree of overlap at the chosen threshold, which is more closely related to the Gscore.

\begin{figure*}[t]
  \centering
  \includegraphics[width=1.\linewidth]{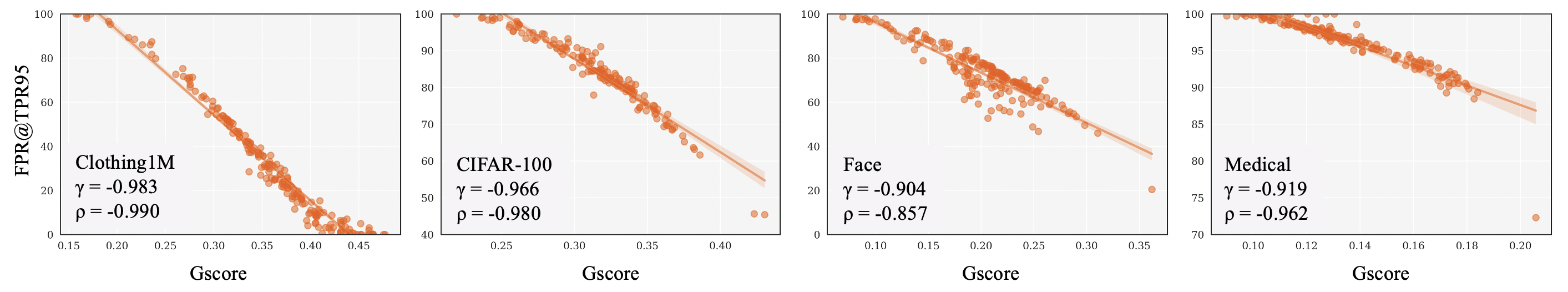}
    \caption{Strong correlation between Gscore and OOD detection performance still exists under various IND datasets. OOD detector is MSP. When a certain IND dataset is used, we remove datasets in Gbench that have overlapping classes with the IND dataset.}
      \vspace{-4mm}
    \label{Clothing1M}
\end{figure*}

\subsection{Variant studies and analysis}
\label{analysis}

\begin{table}[]
\centering
\caption{Unsupervised evaluation with different backbones. Acc represents the classification accuracy (\%) of the backbone on the IND dataset. LT represents latent truth, which is the mean value of the FPR@TPR95 (\%) of the 50 test sets. Pred is the mean value of the predicted FPR@TPR95 of the 50 test sets. Our proposed method predicts very accurate detection performance without OOD labels. We also observe that the classification performance (higher the better) of the backbone shows a positive correlation with its OOD detection performance (lower the better).}
\label{Backbone}
\resizebox{\linewidth}{!}{
\begin{tabular}{lcccc}
\Xhline{1.2pt}
Backbone  & VGG   & ResNet-18 & DenseNet-121 & ResNet-101 \\ \hline
Acc  & 91.06 & 93.59     & 95.50        & 95.68      \\ \hline
LT & 73.15 & 61.44     & 53.29        & 51.08      \\ 
Pred & 75.11 & 60.31     & 52.98        & 49.94      \\ 
\Xhline{1.2pt}
\end{tabular}}
\end{table}

\textbf{Different backbones and IND sets.}
We conduct experiments with different backbones, VGG \cite{simonyan2014very}, ResNet \cite{he2016deep} and DenseNet \cite{huang2017densely}. The results on 150 meta-train sets are shown in Fig.~\ref{different backbones}, which indicates the strong correlation between Gscore and OOD detection performance holds across different backbones, with the absolute value of $\gamma$ consistently higher than 0.952. We further evaluate our method on 50 test sets, and the result is shown in Table~\ref{Backbone}. It indicates that our method accurately predicts the OOD detection performance on unlabelled test sets, which means our method generalizes well on different backbones. We also observe that the classification accuracy of the backbone shows a positive correlation with the OOD detection performance, which is consistent with previous findings \cite{vaze2022openset, hendrycks2019using}.

\textbf{Impact of different IND data.}
To further analyze our performance prediction method, we change the IND dataset to Clothing1M \cite{xiao2015learning}, Face images \cite{goodfellow2013challenges}, CIFAR-100 \cite{krizhevsky2009learning}, and Medical images \cite{cell}. We plot FPR@TPR95 against Gscore in Fig.~\ref{Clothing1M} with the MSP detector. We notice for medical images, there is an OOD dataset point with very good performance. We find that dataset contains various colormaps while medical images are grayscale images, which makes OOD detection very easy. More analysis and experiments with other detectors are shown in the supplementary material. The strong correlation between Gscore with FPR@TPR95 can still be observed in Fig.~\ref{Clothing1M}, which indicates our method works well under different IND datasets. 

\begin{figure*}[t]
  \centering
  \includegraphics[width=1.\linewidth]{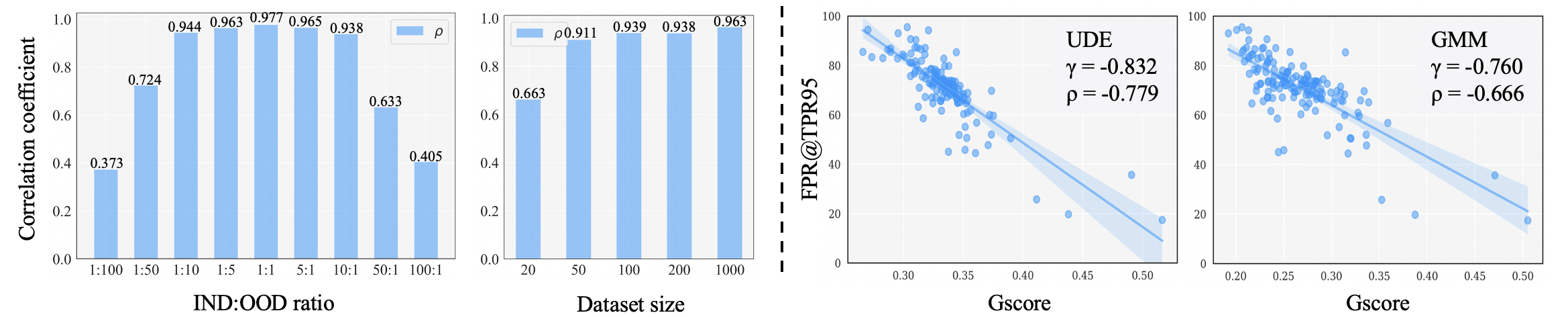}
    \caption{Impact of the test set IND:OOD ratio and size on unsupervised evaluation. We find UDE robust to IND:OOD ratios and dataset sizes. On the right, we create meta-train sets with various IND:OOD ratios and sizes, where UDE shows stronger correlation than GMM.}
    \vspace{-3mm}
    \label{size}
\end{figure*}

\begin{table}[]
\centering
\caption{Unsupervised evaluation of OOD detection methods with different IND datasets. Acc, LT and Pred are of the same meaning as in Table \ref{Backbone}. Our proposed method generalizes well when IND datasets are different. We find the OOD detection performance varies significantly when IND datasets are different.}
\label{IND}
\resizebox{\linewidth}{!}{
\begin{tabular}{lcccc}
\Xhline{1.2pt}
IND set                                                   & Clothing1M & Face & CIFAR-100 & Medical \\ \hline
Acc & 71.26      & 73.25       & 71.36     & 70.83          \\ \hline

LT     & 29.74      & 74.84      & 86.60     & 95.59          \\
Pred   & 29.15     & 74.92       & 88.31     & 97.00          \\ \Xhline{1.2pt}
\end{tabular}}
\vspace{-4mm}
\end{table}

Moreover, we find from Table~\ref{IND} that IND datasets have a non-negligible influence on the OOD detection performance. While the backbones have similar classification accuracy on IND datasets, the OOD detection performance varies in a wide range. There might be a few possible reasons. For example, Clothing1M, compared with CIFAR-100 may be much more different from the OOD datasets. It is also possible that Clothing1M forms a very compact feature space due to its small inter-class distance, and thus makes other label spaces more distinguishable. The difficulty of face images lies between Clothing1M and CIFAR-100. The detection performance is low when IND data are medical images. We speculate the reason lies in that the medical image dataset has only two classes, which impedes the backbone learning to extract useful features to generalize to various OOD datasets. Under all different IND datasets, our method generalizes well and predicts accurate OOD detection performance compared with the latent truth.

\begin{figure*}[t]
  \centering
  \includegraphics[width=1.\linewidth]{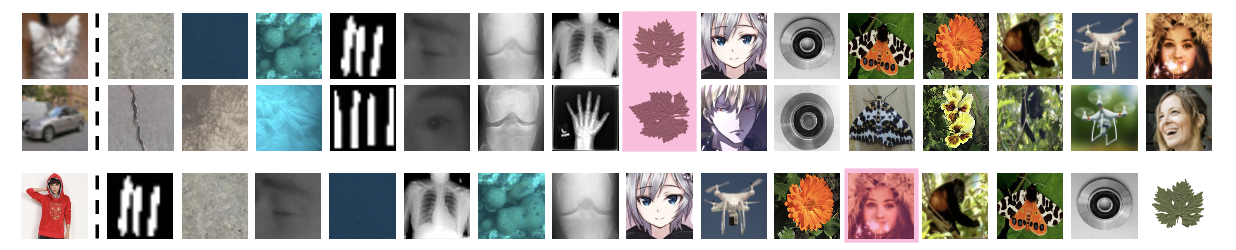}
    \caption{Visualization of the difficulty levels of same OOD datasets when the IND data are CIFAR-10 and Clothing1M, respectively. The first column is the IND data, \textit{i.e.}, CIFAR-10 and Clothing1M. \emph{From left to right, the OOD data are listed from easy to hard.} The red shade marks the most difficult OOD data with the other IND data. With different IND data, the difficulty of the same OOD data is very different.}
    \label{different OOD}
    \vspace{-3mm}
\end{figure*}

\textbf{Impact of the size and IND:OOD ratio of test sets.} In OOD detection community, it is typically assumed that test sets are reasonably large and have a similar number of IND and OOD test samples. However, both aspects might vary in practice. We thus analyze whether our method is still effective under different scenarios. To change the IND:OOD ratio,   
we first make sure each test set has IND:OOD ratio of 1:1, we then down-sample IND samples to make the ratio smaller, while down-sample OOD samples to make the ratio larger, generating a ratio from 1:100 to 100:1. To change the size of the test set, we keep the IND:OOD ratio at 1:1, and reduce the total sample number. Results are summarized in Fig. \ref{size}, from which we have the following findings.

\textbf{First}, when the ratio of IND to OOD data increases from 1:100 to 1:1 and then to 100:1, the correlation coefficient $\rho$ first increases and then decreases. The correlation remains at a high level between 1:10 and 10:1, which demonstrates the robustness of our algorithm. 

\textbf{Second}, we find our method effective when the test set contains as few as 50 or 100 samples, effectively meaning mini-batches. It means we could predict OOD detection performance with mini-batches instead of a large test set. 

\textbf{Third}, we further carry out experiments on Gbench with both different ratios of IND and OOD data and different dataset sizes, which creates a rigorous evaluation scenario. The dataset size generation details are provided in the supplementary material. The results are shown in the right of Fig. \ref{size}, UDE outperforms GMM with the absolute value of correlation coefficient $\gamma$ as 0.832 versus 0.760. The reason lies in that GMM neglects the IND distribution in the original meta-train set, which degrades its performance when the dataset sizes of IND and OOD data are highly imbalanced.

\textbf{Predicting other OOD detection performance metrics.}
We demonstrate that our proposed method can predict various OOD detection metrics. We report the correlation coefficient $\gamma$ and $\rho$. The results in Tabele \ref{metric} show that Gscore generalizes well to other metrics. The performance on F@T60 is low because when TPR=60\%, FPR is close to 0, in such case, the strong correlation ends with a flat line.

\textbf{Difficulty of OOD sets: preliminary analysis.} The Gbench allows us not only to evaluate the accuracy prediction performance, but also to visualize and analyze the difficulty of different datasets in OOD detection, \textit{i.e.}, which types of OOD data are harder to detect \textit{w.r.t} a given IND set? To answer this question, we use different IND datasets and visualize OOD datasets from easy to hard in 
Fig.~\ref{different OOD}. The names of all the datasets are in the supplementary material.

\begin{table}[t]
\centering
\caption{Feasibility of predicting other OOD detection metrics. DE denotes detection error, and F@T means FPR@TPR.}
\label{metric}
\resizebox{\linewidth}{!}{
\begin{tabular}{lcccccc}
\Xhline{1.2pt}
Metric   & DE       & F@T95  & F@T80  & F@T60 & AUROC & AUPR  \\ \hline
Pearson  & -0.976 &   -0.977 & -0.866 & -0.746 &0.842 & 0.771\\
Spearman & -0.968 &   -0.971 & -0.923 & -0.826 &0.900 & 0.794 \\ \Xhline{1.2pt}
\end{tabular}
}
\vspace{-6mm}
\end{table}

When CIFAR-10 is used as IND data, the most difficult OOD datasets are those related to faces, drones and monkeys. We speculate that faces and monkeys and their background look like the \textit{cat} and \textit{dog} categories in CIFAR-10. Drones have similar appearance to \textit{airplane}. When we switch to Clothing1M as IND data, the hard OOD datasets are completely different. We notice Clothing1M contains many images with white ground, which makes OOD datasets with white backgrounds harder. Interestingly, the difficulty of OOD datasets changes drastically under different IND datasets. For example, the leaves dataset is a very hard OOD dataset for Clothing1M, while it is easy for CIFAR-10. We also notice that under both IND datasets, easy OOD datasets remain more or less the same: they generally have simple colors and simple patterns, such as crack images and number images. The above visualization and analysis are still preliminary but very interesting. We will investigate into this dataset understanding problem by developing more principled tools. A future direction would be tailoring OOD detectors for specific IND datasets.

\section{Conclusion}
\label{conclusion}
When we deploy an OOD detection method, from the unlabelled test samples we can observe a distribution of OOD scores. From this distribution only, we aim to predict how well the OOD detector performs. We are particularly interested in this unsupervised evaluation problem under various OOD datasets with a wide range of label spaces. We design a Gscore indicator, computed as the distribution difference between IND and OOD samples, which can be best modeled by our proposed unilateral density estimation approach. On a newly collected benchmark with 200 datasets of various label spaces, we show that Gscore exhibits a strong correlation with OOD detection performance and allows us to use linear regression to accurately predict the performance without OOD labels. We show our algorithm is stable under various OOD detectors, backbones and IND datasets. We also provide interesting insights of the effect of different backbones, IND datasets, and the difficulty of OOD datasets.

{\small
\bibliographystyle{ieee_fullname}
\bibliography{egpaper_for_review.bbl}
}

\clearpage
\appendix
\section{The AUROC of Datasets in Gbench}
\begin{figure}[b]
  \centering
  \includegraphics[width=1.\linewidth]{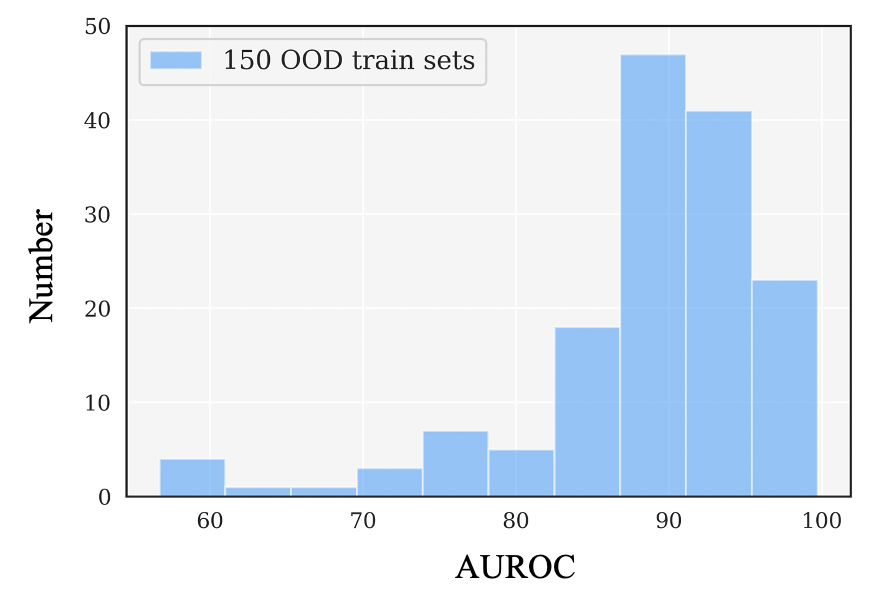}
    \caption{The AUROC distribution of the 150 OOD train sets in Gbench. The wide range of AUROC distribution enables us to train a regression model to predict the AUROC of unlabeled OOD test sets.}
    \label{hist}
\end{figure}

Our collected dataset suite Gbench has various OOD datasets with different difficulties. We present the AUROC distribution of 150 OOD train sets (150 OOD sets mixed with CIFAR-10 test set separately) in Fig. \ref{hist}. The ODIN method is used for detection. We observe that the AUROC ranges from around 57\% to 100\% (random guess yields AUROC of 50\%), indicating that datasets in Gbench have a wide span in their OOD difficulty.

\begin{figure*}[t]
  \centering
  \includegraphics[width=1.\linewidth]{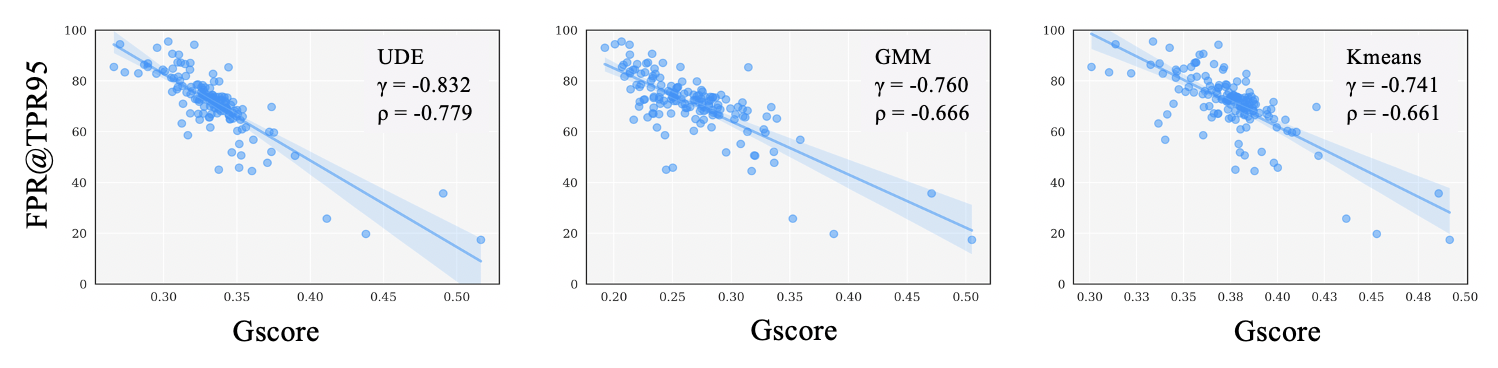}
    \caption{The effect of IND:OOD ratio and dataset size to unsupervised evaluation methods. We evaluate UDE, GMM, Kmeans with train sets of different IND:OOD ratios and dataset sizes. The correlation coefficient indicates that UDE performs the best.}
    \label{size}
\end{figure*}

\section{The Hyperparameter Tuning Process}
We choose the hyperparameter $\tau$ by enumerating its possible values from $[0, 1]$ in the 150 OOD train sets. Our target is to minimize the regression loss on the 150 OOD train sets. Thus, we first enumerate $\tau$ with an interval of $0.1 (0, 0.1, 0.2, ..., 1.0)$, we record the regression loss on each selected $\tau$. After selecting the initial $\tau$ in the interval, for example, $[0.9, 1.0]$, we further enumerate the $\tau$ with an interval of $0.01$ and choose the $\tau$ with the minimal regression loss. We can also select the hyperparameter by enumerating $\tau$ from $[0, 1]$ with an interval of $0.01$. When IND:OOD ratio is 1:1, the $\tau$ is set to 0.99 using UDE, 0.95 using GMM. When the IND:OOD ratio is different across the OOD train sets, the $\tau$ is set to 0.93 using UDE and 0.50 using GMM. GMM and UDE are not very sensitive to the hyperparameter $\tau$, thus, we tune the hyperparameter with OOD detector MSP and use the tuned hyperparameter across other OOD detectors. 

\begin{table*}[t]
\begin{center}
\setlength{\tabcolsep}{0.9mm}

\begin{tabular}{l|ll|cccccc|c}
\Xhline{1.2pt}
\begin{tabular}[c]{@{}l@{}}OOD\\ detectors\end{tabular} & \multicolumn{2}{l|}{\begin{tabular}[c]{@{}l@{}}Unsupervised\\ evaluation method\end{tabular}}                          & Ischemia    & Watch       & Potato disease & Bengali     & Vegetable   & Men women   & RMSE \\ \hline
\multirow{3}{*}{MSP}                                           & \multicolumn{2}{l|}{Kmeans + $\ell 2$}                                                                                 & 83.54/76.66 & 60.97/66.87 & 69.56/77.38    & 82.87/80.10 & 75.13/72.83 & 75.36/70.01 & 5.55 \\

                                                               & \multicolumn{2}{l|}{GMM + Wasserstein}                                                                       & 83.54/84.30 & 60.97/64.77 & 69.56/72.70    & 82.87/84.13 & 75.13/73.71 & 75.36/74.28 & 2.22 \\ 
                                                               
                                                               & \multicolumn{2}{l|}{UDE + Wasserstein}                                                                        & 83.54/84.82 & 60.97/63.39 & 69.56/69.54    & 82.87/82.40 & 75.13/73.83 & 75.36/74.06 & \textbf{1.36} \\ \hline
\multirow{3}{*}{ODIN}                                          & \multicolumn{2}{l|}{Kmeans + $\ell 2$}                                                                                 & 26.82/40.27 & 30.04/30.40 & 52.38/53.00    & 50.87/42.56 & 50.92/47.27 & 27.73/27.71 & 6.63 \\
                                                               & \multicolumn{2}{l|}{GMM + Wasserstein}                                                                       & 26.82/42.08 & 30.04/32.85 & 52.38/64.42    & 50.87/46.40 & 50.92/47.50 & 27.73/30.04 & 8.39 \\

                                                               & \multicolumn{2}{l|}{UDE + Wasserstein}                                                                         & 26.82/37.52 & 30.04/30.64 & 52.38/58.21    & 50.87/51.69 & 50.92/49.72 & 27.73/28.17 & \textbf{5.02} \\ \hline
\multirow{3}{*}{ENERGY}                        & \multicolumn{2}{l|}{Kmeans + $\ell 2$}                                                                                 & 53.61/55.14 & 24.88/32.54 & 45.14/56.46    & 67.91/59.04 & 53.97/50.11 & 49.18/47.03 & 6.92 \\                
& \multicolumn{2}{l|}{GMM + Wasserstein}                                                                       & 53.61/57.80 & 24.88/32.31 & 45.14/54.41    & 67.91/56.56 & 53.97/51.09 & 49.18/48.53 & 7.03 \\

                                                               & \multicolumn{2}{l|}{UDE + Wasserstein}                                                                     & 53.61/53.58 & 24.88/29.38 & 45.14/52.29    & 67.91/65.28 & 53.97/52.65 & 49.18/47.21 & \textbf{3.74} \\ \hline
\multirow{3}{*}{MLS}                                         & \multicolumn{2}{l|}{Kmeans + $\ell 2$}                                                                                 & 56.12/57.09 & 27.40/33.48 & 45.19/56.48    & 67.96/60.13 & 54.57/50.84 & 49.66/48.18 & 6.36 \\ 

& \multicolumn{2}{l|}{GMM + Wasserstein}                                                                       & 56.12/65.96 & 27.40/32.81 & 45.19/53.04    & 67.96/54.97 & 54.57/50.02 & 49.66/54.68 & 8.19 \\

                                                               & \multicolumn{2}{l|}{UDE + Wasserstein}                                                                         & 56.12/55.64 & 27.40/30.65 & 45.19/51.94    & 67.96/65.94 & 54.57/53.41 & 49.66/48.21 & \textbf{3.26} \\ \Xhline{1.2pt}
\end{tabular}
\end{center}
\caption{Evaluation of OOD unsupervised evaluation methods on predicting FPR@TPR95 of unlabeled test sets. The results are shown in the form of latent truth/predicted performance (\%). We evaluate their performance using RMSE (\%), which is smaller when the predicted performance is closer to the latent truth. Ischemia, Watch, Potato disease, Bengali, Vegetable, Men women are six randomly selected unlabeled OOD test sets. UDE + Wasserstein achieves the best performance under all four different OOD detection methods.}
\label{autoeval11}
\end{table*}

\section{Influence of Sizes of IND and OOD Data}
To explore the influence of sizes of IND and OOD data on UDE and other unsupervised evaluation methods, we generate 150 OOD train sets of different sizes of IND and OOD data. Specifically, we randomly select a number from $[100, N]$ as the size of the IND or OOD data, where $N$ is the corresponding size of IND or OOD data. The IND and OOD size of the 150 OOD train sets are listed below:

IND: [6990, 5966, 8061, 2381, 4204, 5181, 7835, 7213, 7915, 1120, 6634, 8186, 5428, 3732, 7439, 8422, 9131, 5551, 9065, 1601, 9531, 3202, 7907, 2550, 8957, 8694, 3625, 4609, 1449, 9718, 365, 3712, 7081, 3684, 8852, 2138, 6509, 696, 2129, 473, 4358, 5032, 1100, 9872, 4362, 2875, 2692, 8774, 7346, 6815, 5951, 9284, 5603, 2309, 4816, 5183, 1422, 5578, 7442, 616, 866, 4347, 8083, 5440, 6901, 2238, 5280, 138, 1702, 5054, 7892, 4600, 2285, 2630, 766, 5256, 788, 7234, 3505, 245, 5562, 5641, 2784, 6013, 4907, 890, 1261, 5002, 9288, 5730, 8254, 5589, 6253, 3344, 228, 8603, 8105, 3659, 3706, 7165, 1146, 7499, 6620, 3637, 7610, 2579, 2531, 7034, 8266, 3684, 5314, 2530, 6309, 7807, 745, 765, 7449, 7648, 8780, 8586, 4845, 8000, 435, 8404, 1175, 6785, 2783, 4803, 7986, 9962, 6836, 7618, 2580, 5406, 3294, 1085, 2136, 6451, 3198, 6187, 1334, 4284, 9426, 1631, 6983, 2486, 1476, 1725, 7419, 8236]

OOD: [3255, 1076, 2194, 8368, 1653, 2507, 2804, 5896, 9152, 2233, 864, 5112, 4096, 2424, 378, 5343, 5038, 2342, 5041, 7390, 5294, 3112, 4360, 2233, 1939, 23210, 3958, 2474, 3027, 1096, 4081, 1544, 2824, 249, 2697, 537, 2061, 2001, 3130, 1100, 515, 1279, 3054, 745, 1733, 9435, 1715, 801, 9877, 7827, 5202, 2613, 747, 1250, 2985, 2268, 13680, 3250, 1927, 6889, 54917, 1146, 4421, 8203, 304, 1467, 3520, 1973, 5114, 3352, 1741, 457, 4302, 1042, 446, 18396, 7058, 7614, 465, 2484, 2323, 1635, 1204, 692, 2224, 4027, 4384, 5482, 1728, 2861, 337, 1336, 9348, 1487, 3252, 9217, 7637, 831, 6134, 2829, 15174, 8714, 4172, 789, 3343, 9052, 2098, 1407, 5383, 2323, 5880, 8703, 2654, 1570, 1157, 634, 24967, 1319, 491, 842, 1788, 6945, 3926, 719, 2825, 1340, 355, 4014, 688, 8158, 5879, 12770, 7815, 1187, 5779, 1918, 1432, 1141, 4351, 7383, 5941, 8065, 2830, 570, 8680, 332, 3557, 1130, 603, 593]

The IND:OOD ratio and the dataset size of different OOD train sets are all different. For example, OOD train set 1 has 6990 IND samples and 3255 OOD samples (IND:OOD ratio is 2.15, dataset size is 10245), while OOD train set 2 has 5966 IND samples and 1076 OOD samples (IND:OOD ratio is 5.54, dataset size is 7042).

We carry out experiments on the 150 generated OOD train sets to evaluate the performance of different unsupervised evaluation methods. The experiment results are shown in Fig. \ref{size}. UDE achieves the best performance as the absolute value of Pearson Correlation Coefficient $\gamma$ reaches 0.832, outperforming 0.760 and 0.741.

\begin{figure}[t]
  \centering
  \includegraphics[width=1.\linewidth]{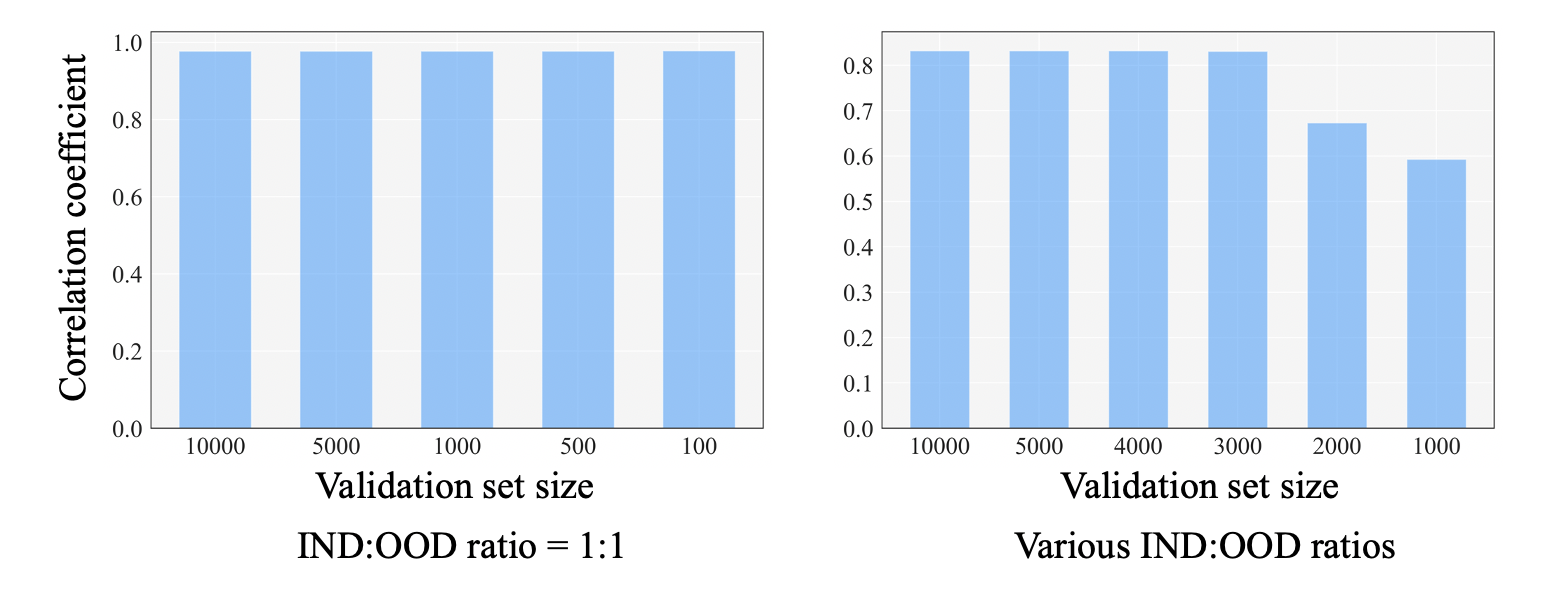}
    \caption{The effect of the validation set size on the performance of OOD unsupervised evaluation methods. The validation set size has little influence on our method when the IND:OOD ratio of train sets is 1:1, as the absolute values of correlation coefficient $\gamma$ remain high ($>$0.97). When the train sets have various IND:OOD ratios, we need around 3000 samples in the validation set to keep our method effective.}
    \label{validation}
\end{figure}

\begin{table*}[t]
\begin{center}
\setlength{\tabcolsep}{0.6mm}
\begin{tabular}{ll|cccc|cccc}
\Xhline{1.2pt}
\multicolumn{2}{l|}{}                           & \multicolumn{4}{c|}{FPR@TPR95}                 & \multicolumn{4}{c}{AUROC}                     \\ \cline{3-10} 
\multicolumn{2}{l|}{Unsup. Eval. Methods}       & MSP       & ODIN       & ENERGY    & MLS       & MSP       & ODIN      & ENERGY    & MLS       \\ \hline
\multicolumn{2}{l|}{Gscore (Kmeans + l2)}       & 8.73±1.22 & 10.97±0.86 & 6.79±0.73 & 6.93±0.77 & 5.32±0.75 & 5.25±0.38 & 5.01±0.75  & 5.17±0.79 \\
\multicolumn{2}{l|}{Gscore (GMM + KL Divergence)} & 5.19±0.56 & 13.44±0.94  & 7.10±0.62 & 7.40±0.73 & 4.24±0.66 & 7.18±0.97 & 6.75±0.76 & 6.32±0.99 \\ 
\multicolumn{2}{l|}{Gscore (GMM + Wasserstein)} & 4.41±0.66 & 6.89±0.46  & 7.28±0.79 & 6.30±0.68 & 4.81±0.55 & 4.53±0.41 & 5.44±0.84 & 4.85±0.54 \\ 
\multicolumn{2}{l|}{Gscore (UDE + KL Divergence)} & 4.10±0.92 & 4.77±1.08  & 4.67±1.30 & 5.07±1.34 & 3.90±0.38 & \textbf{3.80±0.61} & \textbf{3.92±0.65} & \textbf{4.04±0.62} \\ 
\multicolumn{2}{l|}{Gscore (UDE + Wasserstein)} & \textbf{3.46±0.63} & \textbf{4.50±0.23}  & \textbf{4.39±0.55} & \textbf{4.52±0.57} & \textbf{3.64±0.27} & 3.86±0.32 & 4.02±0.51 & 4.08±0.52 \\ \Xhline{1.2pt}
\end{tabular}
\end{center}
\caption{Comparing Gscore variants in unsupervised evaluation of OOD detection. 
We use four OOD detectors and the DLA backbone and predict FPR@TPR95 and AUROC. Each train/test split on Gbench is repeated 5 times, and the mean and standard deviation of RMSE (\%) are reported; the lower the better. We observe UDE and Wasserstein distance achieves the best performance overall.}
\label{autoeval}
\end{table*}

\section{Examples of Unsupervised Evaluation}
We show some examples of unsupervised evaluation on unlabeled test sets in Table \ref{autoeval11}. We train the regression model on all 200 datasets of Gbench and collect another 6 unlabeled OOD sets for testing. Ischemia contains ECGs images of patients with Ischemia. Watch contains watch images. Potato disease contains potato leaves images for potato disease detection. Bengal contains sign language images. Vegetable contains images of vegetables. Men women contains images of men and women. We display the latent truth OOD detection performance and the predicted OOD detection performance. RMSE is calculated on the 6 unlabeled test sets for each method. The results show that UDE+Wasserstein (Ours) achieves the best performance under all four different OOD detectors. The best performance is achieved when the OOD detector is MSP, the RMSE is only 1.36\%.

\section{Validation Set Size on UDE Performance}
Our proposed UDE needs a validation set to infer the distribution of OOD scores of the IND data. We show the validation set size has negligible influence on the performance of our method in Fig. \ref{validation}. When the OOD train sets have IND:OOD ratio of 1:1, then the validation set size has little effect on the performance of our unsupervised evaluation method. With only the validation set size of 100, we can find the strong correlation between Gscore and the OOD detection performance as $\gamma$ is over 0.97. When the OOD train sets have various IND:OOD ratios and dataset sizes, validation set size has a relatively stronger influence. We need around 3000 validation samples to keep the unsupervised evaluation method effective.

\section{OOD Sets with Different IND Sets}
\label{numberood}
When the IND dataset is changed, we need to remove the OOD datasets that have class overlap with the new IND dataset. When the IND dataset is Clothing1M, we remove Store, Fashion and Shoes sandals boots, and there remain 197 OOD datasets in Gbench. When the IND dataset is Face, we remove Faces age detection, RAF-DB, Affectnet, FER2013, Gender, LFW, Good and bad guys, Face mask, Female and male eyes, Real and fake face, Fake video and there remain 189 OOD datasets. When the IND dataset is Medical images, we remove Brain tumor, Medical MNIST, Breast ultrasound, Skin cancer MNIST, Knee X-Ray, Alzheimer MRI, Brain tumor MRI, Br35H, ChestX, COVID19, Brain CT, Kvasir and there remain 188 OOD datasets. When the IND dataset is CIFAR-100, we remove Faces age detection, MIML, Bonsai styles images, MIT indoor scenes, 102flowers, iSUN, RAF-DB, Affectnet, Sea animals, Facial expression, Fish, Gender, Butterfly, Imagenet, Monkey, ImageNet-A, ImageNet-O, ImageNet-R, Tiny-ImageNet resize, Intel image, Elephant, Snake, LFW, Flower, Tin and steel cans, Mobile and smartphone, Beauty classification, Good and bad guys, Plastic and paper cups, Cloud, Furniture, Natural scene, Flower extension, Ornamental plants, Butterflies 100, Weather, Real and fake face, Insect village synthetic, Blossom, AFFiNe, Fight, Fish species, Split garbage, Fruit and vegetable, Wild plants, Pests, Cattle, Fake video, Apple, Toxic plant, Pistol detection, Clock, Crocodile and there remain 147 OOD datasets.

\begin{figure*}[t]
  \centering
  \includegraphics[width=1.\linewidth]{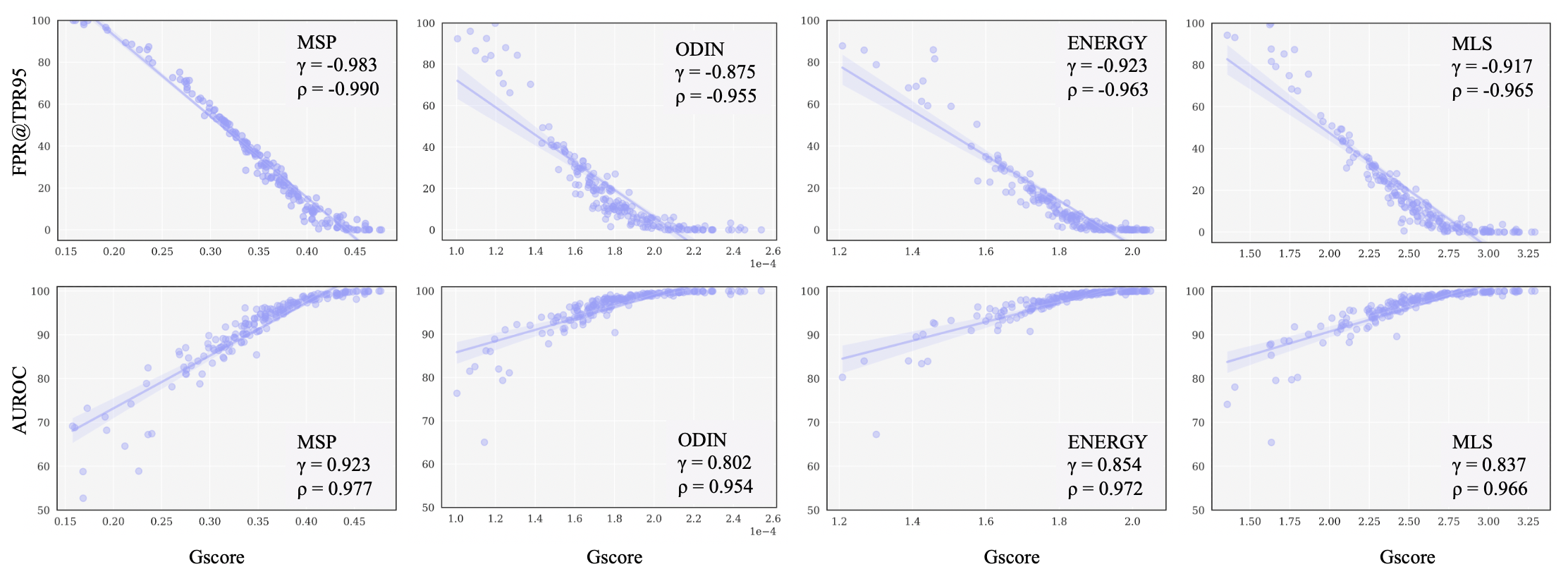}
    \caption{The correlation between Gscore and OOD detection performance under different detectors. The IND dataset is Clothing1M. We can observe the strong correlation still exists under different detectors. The correlation becomes flat when Gscore is large enough, which can be attributed to the strong performance of OOD detectors on some relatively easy test sets, where FPR@TPR95 and AUROC reach saturation (close to 0 or 100)}
    \label{ind}
\end{figure*}

\section{More Results of Other Distance Metrics}

We add results of UDE and GMM with KL Divergence in Table \ref{autoeval} compared with the results in the paper. Kmeans utilizes $\ell 2$ loss as Kmeans only returns the centers of the two distributions. GMM and UDE do not use $\ell 2$ loss as they fit the OOD scores into two Gaussian distributions, we further consider the variances of the Gaussian distributions as different variances under the same mean values might lead to different OOD detection performance. Shown in Table \ref{autoeval}, we can observe UDE+Wasserstein is the most effective unsupervised evaluation method on different OOD detectors. UDE+KL Divergence is slightly weaker than UDE+Wasserstein. GMM works well with Wasserstein distance while GMM + KL Divergence has low performance under ODIN, ENERGY and MLS. We speculate the reason lies in that the standard deviance is not accurately estimated by GMM. For example, for a random OOD train set, the estimated $\mu_{1}$, $\mu_{2}$, $\sigma_{1}$, $\sigma_{2}$ by GMM are 0.1009, 0.1006, $8.85e-5$,  $1.05e-4$, while the latent truths are 0.1009, 0.1007, $1.15e-4$, $1.7e-4$, we can see the estimated mean values of GMM are close to latent truth, while the estimated standard deviance values are much more different from latent truth. We further utilize $\ell$ 2 loss to only compute the distance between the estimated $\mu_{1}$ and $\mu_{2}$ by GMM, the RMSE is 9.77 under MLS detector with seed as 10000, when we utilize KL Divergence loss to incorporate the estimated $\sigma_{1}$ and $\sigma_{2}$, the RMSE is 18.09, which further validate our speculation that the performance is low because the estimated standard deviance of GMM is not accurate.

\begin{figure}[]
  \centering
  \includegraphics[width=1.\linewidth]{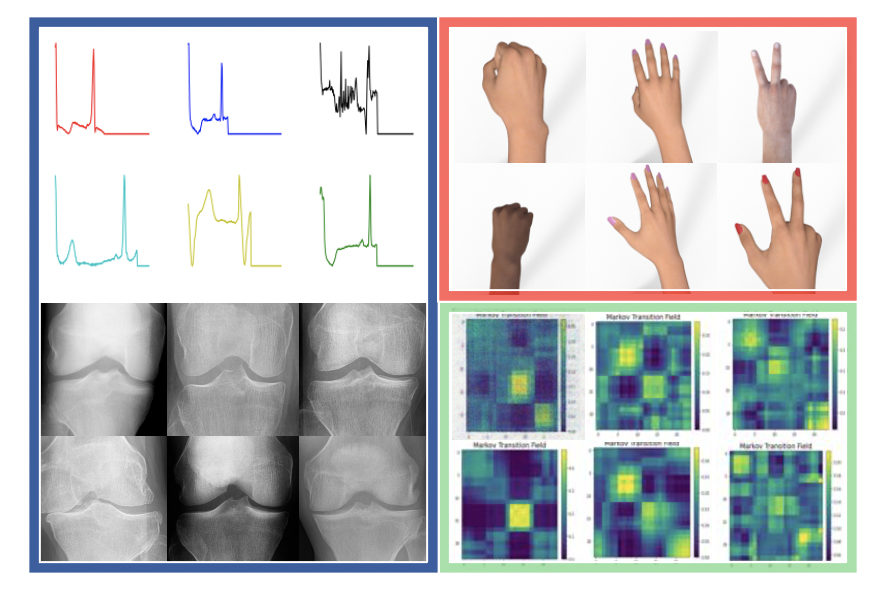}
    \caption{The outlier OOD datasets with different IND datasets. Blue represents that CIFAR-100 is the IND dataset, red represents that Face is the IND dataset, green represents that Medical is the IND dataset.}
    \label{outlier}
\end{figure}

\section{Results on Other OOD Detectors}

In our paper, we show the experiment results when the IND dataset is changed under detector MSP. We further provide the experiment results in Fig. \ref{ind} on other detectors when the IND dataset is changed to Clothing1M to study the effect of different detectors under different IND datasets. The results illustrate that the strong correlation still exists under different OOD detectors when the IND dataset is changed to Clothing1M. Interestingly, when using Clothing1M as IND data, the correlation becomes flat when Gscore is large enough. This phenomenon can be attributed to the strong performance of OOD detectors on some relatively easy test sets, where FPR@TPR95 and AUROC reach saturation (close to 0 or 100). Under this circumstance, even if Gscore keeps increasing, the OOD detection performance no longer changes.

\section{Visualization of the Outlier OOD Datasets}
We find that when the IND dataset is changed to CIFAR-100, Face or Medical, there are some outlier OOD datasets with very high OOD detection performance. We provide some examples of these datasets in Fig.~\ref{outlier}. When the IND dataset is CIFAR-100, the outlier easiest OOD datasets are ECG signal images and knee X-Ray images, they are with very simple patterns and each image contains almost a single color. When the IND dataset is Face, the easiest OOD dataset is rock paper scissors gestures, we speculate that the main content of gesture images (hands) and face images are with similar color, while the shape of hands is very different from faces plus the background of the two datasets are different, which make the gesture dataset an easy OOD dataset. When Medical is the IND dataset, the easiest OOD dataset is Pump colormaps, the reason might be colormap images have different colors while medical images are mainly gray, which makes the two datasets very different. From the analysis of the outlier datasets of different IND datasets, we further validate our proposal that OOD datasets have different difficulties regards different IND datasets. Future research direction might be studying the difficulty of OOD datasets under different IND datasets and designing specific OOD detectors for different IND datasets.

\section{List of the 6 datasets in Table~\ref{autoeval11}}

\noindent\textbf{Ischemia}:
\url{https://www.kaggle.com/datasets/buraktaci/mri-stroke}

\noindent\textbf{Watch}:
\url{https://www.kaggle.com/datasets/ahedjneed/fancy-watche-images}

\noindent\textbf{Potato}:
\url{https://www.kaggle.com/datasets/rizwan123456789/potato-disease-leaf-datasetpld}

\noindent\textbf{Bengali}:
\url{https://www.kaggle.com/datasets/muntakimrafi/bengali-sign-language-dataset}

\noindent\textbf{Vegetable}:
\url{https://www.kaggle.com/datasets/misrakahmed/vegetable-image-dataset}

\noindent\textbf{Menwomen}: \url{https://www.kaggle.com/datasets/playlist/men-women-classification}

\section{Complete list of the 200 datasets in Gbench}

\noindent\textbf{Rock} : Different Types of Rocks Images
\url{https://www.kaggle.com/datasets/shlokjain69/rock-classification}

\noindent\textbf{Faces age detection} : Predict the Age of an Actor or Actress from Facial Attributes
\url{https://www.kaggle.com/datasets/arashnic/faces-age-detection-dataset}

\noindent\textbf{MIML} : Multi-instance Multi-label Classification under Natural Scene
\url{https://www.kaggle.com/datasets/twopothead/miml-image-data}

\noindent\textbf{Tally marks} : Tally Marks Image Dataset
\url{https://www.kaggle.com/datasets/duraidkhaalid/tally-marks}

\noindent\textbf{Color polygon images} : The Simplest Dataset for Classification and Regression Practice
\url{https://www.kaggle.com/datasets/gonzalorecioc/color-polygon-images}

\noindent\textbf{Bonsai styles images} : 2700 Images of Bonsai's Styles, 9 Classes
\url{https://www.kaggle.com/datasets/vincenzors8/bonsai-styles-images}

\noindent\textbf{Drowsiness detection} : Human Eye Images
\url{https://www.kaggle.com/datasets/kutaykutlu/drowsiness-detection}

\noindent\textbf{MIT indoor scenes} : Indoor Scene Recognition
\url{https://www.kaggle.com/datasets/itsahmad/indoor-scenes-cvpr-2019}

\noindent\textbf{102flowers} : Flower Images Dataset for Classification with a Large Number of Classes
\url{https://www.kaggle.com/datasets/hishamkhdair/102flowers-data}

\noindent\textbf{Eurosat} : Dataset Contains All the RGB and Bands Images from Sentinel-2
\url{https://www.kaggle.com/datasets/apollo2506/eurosat-dataset}

\noindent\textbf{iSUN} : A Saliency Dataset for A Large Number of Natural Images
\url{https://turkergaze.cs.princeton.edu/}

\noindent\textbf{Santa} : Santa Claus Classification
\url{https://www.kaggle.com/datasets/deepcontractor/is-that-santa-image-classification}

\noindent\textbf{RAF-DB} : Real-world Affective Faces Database (RAF-DB)~\cite{li2017reliable}
\url{http://www.whdeng.cn/raf/model1.html}

\noindent\textbf{LSUN} : Large-scale Scene UNderstanding Dataset (LSUN)
\url{https://www.yf.io/p/lsun}

\noindent\textbf{Satellite image} : Satellite Remote Sensing Images
\url{https://www.kaggle.com/datasets/mahmoudreda55/satellite-image-classification}

\noindent\textbf{Affectnet} : Large-scale Facial Expression Recognition Dataset~\cite{mollahosseini2017affectnet}
\url{http://mohammadmahoor.com/affectnet/}

\noindent\textbf{LSUN resize} : Downsampled Version of LSUN.
\url{https://github.com/facebookresearch/odin}

\noindent\textbf{Sea animals} : Images of Different Sea Creatures for Image Classification
\url{https://www.kaggle.com/datasets/vencerlanz09/sea-animals-image-dataste}

\noindent\textbf{Facial expression} : Facial Expression Recognition 2013 Dataset~\cite{goodfellow2013challenges}
\url{https://www.kaggle.com/datasets/msambare/fer2013}

\noindent\textbf{Shells} : A dataset Containing Images of Shells and Pebbles for Image Classification
\url{https://www.kaggle.com/datasets/vencerlanz09/shells-or-pebbles-an-image-classification-dataset}

\noindent\textbf{Alphabet} : Image Dataset for Alphabets in the American Sign Language
\url{https://www.kaggle.com/datasets/grassknoted/asl-alphabet}

\noindent\textbf{Fish} : A Large-Scale Dataset for Fish Segmentation and Classification
\url{https://www.kaggle.com/datasets/crowww/a-large-scale-fish-dataset}

\noindent\textbf{Meat freshness} : Meat Freshness Image Classification Dataset
\url{https://www.kaggle.com/datasets/vinayakshanawad/meat-freshness-image-dataset}

\noindent\textbf{Sign language} : Turkey Sign Language Digits Dataset
\url{https://www.kaggle.com/datasets/ardamavi/sign-language-digits-dataset}

\noindent\textbf{Balls} : 26 Types of Balls - Image Classification
\url{https://www.kaggle.com/datasets/gpiosenka/balls-image-classification}

\noindent\textbf{Food} : Labeled Food Images in 101 Categories from Apple Pies to Waffles
\url{https://www.kaggle.com/datasets/kmader/food41}

\noindent\textbf{Sports} : 100 Sports Image Classification
\url{https://www.kaggle.com/datasets/gpiosenka/sports-classification}

\noindent\textbf{Brain tumor} : Brain Tumor Image Classification
\url{https://www.kaggle.com/datasets/kirolosmedhat264/brain-tumor}

\noindent\textbf{Gender} : Male Female Image Dataset
\url{https://www.kaggle.com/datasets/cashutosh/gender-classification-dataset}

\noindent\textbf{Medical MNIST} : Medical MNIST,
58954 Medical Images of 6 Classes
\url{https://www.kaggle.com/datasets/andrewmvd/medical-mnist}

\noindent\textbf{Breast ultrasound} : Breast Ultrasound Images for Classification, Detection and Segmentation
\url{https://www.kaggle.com/datasets/aryashah2k/breast-ultrasound-images-dataset}

\noindent\textbf{German} : GTSRB - German Traffic Sign Recognition Benchmark
\url{https://www.kaggle.com/datasets/meowmeowmeowmeowmeow/gtsrb-german-traffic-sign}

\noindent\textbf{Messy rooms} : Messy vs Clean Room - A Small Dataset for Scene Image Classification
\url{https://www.kaggle.com/datasets/cdawn1/messy-vs-clean-room}

\noindent\textbf{SVHN} : A Real-world Image Dataset Obtained from House Numbers in Google Street View Images~\cite{Netzer2011ReadingDI}
\url{http://ufldl.stanford.edu/housenumbers/}

\noindent\textbf{Butterfly} : Butterfly Dataset
\url{https://www.kaggle.com/datasets/veeralakrishna/butterfly-dataset}

\noindent\textbf{Grapevine} : Grapevine Leaves Image Dataset
\url{https://www.kaggle.com/datasets/muratkokludataset/grapevine-leaves-image-dataset}

\noindent\textbf{Chinese MNIST
} : Chinese Numbers Handwritten Characters Images
\url{https://www.kaggle.com/datasets/gpreda/chinese-mnist}

\noindent\textbf{ImageNet} : An Image Database Organized According to The WordNet Hierarchy~\cite{5206848}
\url{https://www.image-net.org/update-mar-11-2021.php}

\noindent\textbf{Monkey} : 10 Monkey Species
\url{https://www.kaggle.com/datasets/slothkong/10-monkey-species}

\noindent\textbf{Textures} : Describable Textures Dataset (DTD) - An Evolving Collection of Textural Images in the Wild ~\cite{cimpoi2014describing}
\url{https://www.robots.ox.ac.uk/~vgg/data/dtd/}

\noindent\textbf{ImageNet-A} :  Natural Adversarial Example Dataset for Image Classifiers~\cite{hendrycks2021nae}
\url{https://github.com/hendrycks/natural-adv-examples}

\noindent\textbf{Elephant} : Asian vs African Elephants
\url{https://www.kaggle.com/datasets/vivmankar/asian-vs-african-elephant-image-classification}

\noindent\textbf{ImageNet-O} :  Natural Adversarial Example Dataset for Out-of-distribution Detectors~\cite{hendrycks2021nae}
\url{https://github.com/hendrycks/natural-adv-examples}

\noindent\textbf{Pokemon} : 7000 Hand-Cropped and Labeled Pokemon Images for Classification
\url{https://www.kaggle.com/datasets/lantian773030/pokemonclassification}

\noindent\textbf{Tom and Jerry} : Collection of 5k+ Images with Labelled Data of Tom and Jerry Cartoon Show
\url{https://www.kaggle.com/datasets/balabaskar/tom-and-jerry-image-classification}

\noindent\textbf{ImageNet-R} : Renditions of 200 ImageNet Classes~\cite{hendrycks2021many}
\url{https://github.com/hendrycks/imagenet-r}

\noindent\textbf{Tree nuts} : Tree Nuts Image Classification
\url{https://www.kaggle.com/datasets/gpiosenka/tree-nuts-image-classification}

\noindent\textbf{Tiny-ImageNet resize} : Downsampled Version of Tiny-ImageNet.
\url{https://github.com/facebookresearch/odin}

\noindent\textbf{Instrument} : Collection of Music Instrument Images with Labels
\url{https://www.kaggle.com/datasets/lasaljaywardena/music-instrument-images-dataset}

\noindent\textbf{Intel image} : Intel Image Classification - Image Scene Classification of Multiclass
\url{https://www.kaggle.com/datasets/puneet6060/intel-image-classification}

\noindent\textbf{Recursion cellular} : Recursion Cellular Image Classification
\url{https://www.kaggle.com/datasets/xhlulu/recursion-cellular-image-classification-224-jpg}

\noindent\textbf{Rice} : Five different Rice Image Dataset
\url{https://www.kaggle.com/datasets/muratkokludataset/rice-image-dataset}

\noindent\textbf{Skin cancer MNIST} : A Large Collection of Pigmented Lesions Images
\url{https://www.kaggle.com/datasets/kmader/skin-cancer-mnist-ham10000}

\noindent\textbf{Recycling} : Recycling - Image Classification
\url{https://www.kaggle.com/datasets/aminizahra/recycling2}

\noindent\textbf{Cricket shots} : Augmented Images of 4 Different Cricket Shots
\url{https://www.kaggle.com/datasets/aneesh10/cricket-shot-dataset}

\noindent\textbf{Shoe} : 15,000 Images of Shoes, Sandals and Boots for Classification
\url{https://www.kaggle.com/datasets/hasibalmuzdadid/shoe-vs-sandal-vs-boot-dataset-15k-images}

\noindent\textbf{Domino tiles} : Photographs of 28 Different Domino Tile Classes
\url{https://www.kaggle.com/datasets/bjorkwall/photographs-of-28-different-domino-tiles}

\noindent\textbf{Concrete defect} : Concrete Defect Image Classification
\url{https://www.kaggle.com/datasets/datastrophy/concrete-train-test-split-dataset}

\noindent\textbf{Snake} : HackerEarth Deep Learning Identify The Snake Breed
\url{https://www.kaggle.com/datasets/oossiiris/hackerearth-deep-learning-identify-the-snake-breed}

\noindent\textbf{LFW} : Labelled Faces in the Wild (LFW) Dataset
\url{https://www.kaggle.com/datasets/jessicali9530/lfw-dataset}

\noindent\textbf{Knee X-Ray} : Knee Osteoarthritis Dataset with Severity Grading
\url{https://www.kaggle.com/datasets/shashwatwork/knee-osteoarthritis-dataset-with-severity}

\noindent\textbf{Persian digits} : 30000 Labeled Persian Digits with Noise in the Background.
\url{https://www.kaggle.com/datasets/aliassareh1/persian-digits-captcha}

\noindent\textbf{Four shapes} : 16,000 Images of Four Basic Shapes (Star, Circle, Square, Triangle
\url{https://www.kaggle.com/datasets/smeschke/four-shapes}

\noindent\textbf{BreakHis} : Breast Cancer Histopathological Database (BreakHis)
\url{https://www.kaggle.com/datasets/ambarish/breakhis}

\noindent\textbf{Mechanical tools} : Mechanical Tools Classification Dataset
\url{https://www.kaggle.com/datasets/salmaneunus/mechanical-tools-dataset}

\noindent\textbf{License plate digits} : License Plate Digits and Characters Classification Dataset
\url{https://www.kaggle.com/datasets/aladdinss/license-plate-digits-classification-dataset}

\noindent\textbf{LEGO Bricks} : Images of LEGO Bricks
\url{https://www.kaggle.com/datasets/joosthazelzet/lego-brick-images}

\noindent\textbf{BarkVN-50} : Bark Texture Images Classification
\url{https://www.kaggle.com/datasets/saurabhshahane/barkvn50}

\noindent\textbf{Flower} : 4242 Images of Flowers
\url{https://www.kaggle.com/datasets/alxmamaev/flowers-recognition}

\noindent\textbf{Rating OpenCV} : Rating OpenCV Emotion Images
\url{https://www.kaggle.com/datasets/juniorbueno/rating-opencv-emotion-images}

\noindent\textbf{Tin and steel cans} : 50,000 Synthetic Images of Steel and Tin Cans for Image Classification
\url{https://www.kaggle.com/datasets/vencerlanz09/tin-and-steel-cans-synthetic-image-dataset}

\noindent\textbf{Plastic, paper and garbage bags} : Synthetic Images of Plastic, Paper, and Garbage Bags for Computer Vision Tasks.
\url{https://www.kaggle.com/datasets/vencerlanz09/plastic-paper-garbage-bag-synthetic-images}

\noindent\textbf{Mobile and smartphone} : Collection of Mobile and Smartphone Images with Annotated Labels
\url{https://www.kaggle.com/datasets/lasaljaywardena/mobile-smartphone-images-dataset}

\noindent\textbf{Produce defect} : Casting Product Image Data for Quality Inspection
\url{https://www.kaggle.com/datasets/ravirajsinh45/real-life-industrial-dataset-of-casting-product}

\noindent\textbf{Beauty classification} : Beauty Classification Image Classification
\url{https://www.kaggle.com/datasets/gpiosenka/beauty-detection-data-set}

\noindent\textbf{Crack detection} : Concrete Crack Images for Image Classification
\url{https://www.kaggle.com/datasets/arnavr10880/concrete-crack-images-for-classification}

\noindent\textbf{Mechanical parts} : Images of Mechanical Parts (Bolt,Nut, Washer,Pin)
\url{https://www.kaggle.com/datasets/manikantanrnair/images-of-mechanical-parts-boltnut-washerpin}

\noindent\textbf{Good and bad guys} : Good Guys and Bad Guys Image Dataset
\url{https://www.kaggle.com/datasets/gpiosenka/good-guysbad-guys-image-data-set}

\noindent\textbf{DeepWeedsX} : A Large Weed Species Image Dataset Collected across Northern Australia
\url{https://www.kaggle.com/datasets/coreylammie/deepweedsx}

\noindent\textbf{Plastic and paper cups} : Plastic and Paper Cups Synthetic Image Dataset
\url{https://www.kaggle.com/datasets/vencerlanz09/plastic-and-paper-cups-synthetic-image-dataset}

\noindent\textbf{Alzheimer MRI} : Alzheimer MRI Preprocessed Dataset (Magnetic Resonance Imaging)
\url{https://www.kaggle.com/datasets/sachinkumar413/alzheimer-mri-dataset}

\noindent\textbf{Fashion} : Fashion Product Images
\url{https://www.kaggle.com/datasets/paramaggarwal/fashion-product-images-small}

\noindent\textbf{Brain tumor MRI} : Brain Tumor MRI Dataset
\url{https://www.kaggle.com/datasets/masoudnickparvar/brain-tumor-mri-dataset}

\noindent\textbf{HAR} : Human Action Recognition (HAR) Dataset
\url{https://www.kaggle.com/datasets/meetnagadia/human-action-recognition-har-dataset}

\noindent\textbf{Devanagari} : Devanagari Character Set
\url{https://www.kaggle.com/datasets/rishianand/devanagari-character-set}

\noindent\textbf{One piece} : Manually Selected Images of Some One Piece Characters
\url{https://www.kaggle.com/datasets/ibrahimserouis99/one-piece-image-classifier}

\noindent\textbf{Cloud} : Clouds Images Taken from the Ground
\url{https://www.kaggle.com/datasets/nakendraprasathk/cloud-image-classification-dataset}

\noindent\textbf{Diamond} : Natural Diamonds Dataset
\url{https://www.kaggle.com/datasets/harshitlakhani/natural-diamonds-prices-images}

\noindent\textbf{Utensil} : Binary and Raw Images of 20 Categories of Utensils
\url{https://www.kaggle.com/datasets/jehanbhathena/utensil-image-recognition}

\noindent\textbf{GTZAN} : GTZAN Dataset - Music Genre Classification
\url{https://www.kaggle.com/datasets/andradaolteanu/gtzan-dataset-music-genre-classification}

\noindent\textbf{Trees satellite} : Trees in Satellite Imagery
\url{https://www.kaggle.com/datasets/mcagriaksoy/trees-in-satellite-imagery}

\noindent\textbf{Movie posters} : Movie Posters with Respected Genres
\url{https://www.kaggle.com/datasets/raman77768/movie-classifier}

\noindent\textbf{Furniture} : Collection of Furniture Images Annotated with Labels
\url{https://www.kaggle.com/datasets/lasaljaywardena/furniture-images-dataset}

\noindent\textbf{PepsiCo} : PepsiCo Lab Potato Chips Quality Control Image Dataset
\url{https://www.kaggle.com/datasets/concaption/pepsico-lab-potato-quality-control}

\noindent\textbf{Style color} : Brand and Product Recognition
\url{https://www.kaggle.com/datasets/olgabelitskaya/style-color-images}

\noindent\textbf{Monkeypox skin lesion} : Binary Classification Data for Monkeypox vs Non-monkeypox (Chickenpox, Measles)
\url{https://www.kaggle.com/datasets/nafin59/monkeypox-skin-lesion-dataset}

\noindent\textbf{Crustacea} : Preprocessed Sample Plankton Image Database Containing 24 Classes of Crustacea
\url{https://www.kaggle.com/datasets/iandutoit/crustacea-zooscan-image-database}

\noindent\textbf{Hurricane damage} : Satellite Images of Hurricane Damage
\url{https://www.kaggle.com/datasets/kmader/satellite-images-of-hurricane-damage}

\noindent\textbf{Indian traffic} : Multi-Class Image Classification on Indian Traffic Signs Dataset (85 Classes)
\url{https://www.kaggle.com/datasets/sarangdilipjodh/indian-traffic-signs-prediction85-classes}

\noindent\textbf{B200C} : High Quality Image Classification for the 200 Most Popular LEGO Parts
\url{https://www.kaggle.com/datasets/ronanpickell/b200c-lego-classification-dataset}

\noindent\textbf{Store} : 6000+ Store Items Images Classified by Color
\url{https://www.kaggle.com/datasets/imoore/6000-store-items-images-classified-by-color?select=test}

\noindent\textbf{Crowd counting} : Crowd Counting Dataset
\url{https://www.kaggle.com/datasets/fmena14/crowd-counting}

\noindent\textbf{American sign} : American Sign Language Dataset
\url{https://www.kaggle.com/datasets/ayuraj/asl-dataset}

\noindent\textbf{Natural scene} : Image Classification Dataset of Various Locations
\url{https://www.kaggle.com/datasets/shanmukh05/ml-hackathon}

\noindent\textbf{Treasure} : Museum Art Mediums Image Classification Dataset
\url{https://www.kaggle.com/datasets/ferranpares/mame-dataset}

\noindent\textbf{Br35H} : Brain Tumor Detection 2020
\url{https://www.kaggle.com/datasets/ahmedhamada0/brain-tumor-detection}

\noindent\textbf{Gemstone} : 87 classes of Gemstones for Image Classification
\url{https://www.kaggle.com/datasets/lsind18/gemstones-images}

\noindent\textbf{Flower extension} : Flower Data Extension
\url{https://www.kaggle.com/datasets/eugeneryu/flower-data-extension}

\noindent\textbf{Planets and moons} : Planets and Moons Dataset
\url{https://www.kaggle.com/datasets/emirhanai/planets-and-moons-dataset-ai-in-space}

\noindent\textbf{Rock Paper Scissors} : Images from the Rock-Paper-Scissors Game
\url{https://www.kaggle.com/datasets/drgfreeman/rockpaperscissors}

\noindent\textbf{Ornamental plants} : Image Dataset for Common Flowering Plants
\url{https://www.kaggle.com/datasets/abdalnassir/ornamental-plants}

\noindent\textbf{Coffee bean} : Multiclass Classification Data for Each Seed Roasted Coffee Bean
\url{https://www.kaggle.com/datasets/gpiosenka/coffee-bean-dataset-resized-224-x-224}

\noindent\textbf{Butterflies 100} : 100 Butterfly Species Dataset
\url{https://www.kaggle.com/datasets/gpiosenka/butterflies-100-image-dataset-classification}

\noindent\textbf{FoodyDudy} : The First Ever Database about Thai Food on Kaggle.
\url{https://www.kaggle.com/datasets/somboonthamgemmy/foodydudy}

\noindent\textbf{Tobacco3482} : Document Structure Learning Dataset
\url{https://www.kaggle.com/datasets/patrickaudriaz/tobacco3482jpg}

\noindent\textbf{Artworks} : Collection of Paintings of the 50 Most Influential Artists of All Time
\url{https://www.kaggle.com/datasets/ikarus777/best-artworks-of-all-time}

\noindent\textbf{Face mask} : Face Mask Detection
\url{https://www.kaggle.com/datasets/andrewmvd/face-mask-detection}

\noindent\textbf{Color} : Dataset for Color Classification
\url{https://www.kaggle.com/datasets/ayanzadeh93/color-classification}

\noindent\textbf{Star wars} : Images Classification on Star Wars Characters
\url{https://www.kaggle.com/datasets/mathurinache/star-wars-images}

\noindent\textbf{Weather} : Different Types of Weather Image Dataset
\url{https://www.kaggle.com/datasets/jehanbhathena/weather-dataset}

\noindent\textbf{Real and fake face} : Real and Fake Face Detection
\url{https://www.kaggle.com/datasets/ciplab/real-and-fake-face-detection}

\noindent\textbf{Turkey traffic} : Traffic Sign Images From Turkey
\url{https://www.kaggle.com/datasets/erdicem/traffic-sign-images-from-turkey}

\noindent\textbf{Pattern} : Indonesian Batik Motifs
\url{https://www.kaggle.com/datasets/dionisiusdh/indonesian-batik-motifs}

\noindent\textbf{UAV detection} : UAV Detection Dataset
\url{https://www.kaggle.com/datasets/nelyg8002000/uav-detection-dataset-images}

\noindent\textbf{MIIA pothole} : Images of Roads in South Africa Contain Potholes
\url{https://www.kaggle.com/datasets/salimhammadi07/miia-pothole-image-classification-challenge}

\noindent\textbf{Female and male eyes} : Images of Female and Male Eyes.
\url{https://www.kaggle.com/datasets/pavelbiz/eyes-rtte}

\noindent\textbf{POLLEN20L} : Pollen Dataset of 20 Species Annotated for the Detection
\url{https://www.kaggle.com/datasets/nataliakhanzhina/pollen20ldet}

\noindent\textbf{Coral detection} : Coral Image Classification
\url{https://www.kaggle.com/datasets/bobaaayoung/coral-image-classification}

\noindent\textbf{AmsterTime} : A Visual Place Recognition Benchmark Dataset for Severe Domain Shif t
\url{https://www.kaggle.com/datasets/byildiz/amstertime}

\noindent\textbf{Letters typefaces} : Standard Windows Fonts with Letters Organized in Classes by Typeface
\url{https://www.kaggle.com/datasets/killen/bw-font-typefaces}

\noindent\textbf{Indian dance} : Indian Dance Form Classification
\url{https://www.kaggle.com/datasets/aditya48/indian-dance-form-classification}

\noindent\textbf{Pump colormaps} : Pump Image Classification
\url{https://www.kaggle.com/datasets/byvickey/pump-image-classification}

\noindent\textbf{Komering} : Intelligent System Research Group Bina Darma Dataset
\url{https://www.kaggle.com/datasets/ykunang/aksara-komering}

\noindent\textbf{Simpsons} : Image Dataset of 20 Characters from The Simpsons
\url{https://www.kaggle.com/datasets/alexattia/the-simpsons-characters-dataset}

\noindent\textbf{Cosmos} : A Simple Collection of 3,600 Space Images for Space Image Generation GANs
\url{https://www.kaggle.com/datasets/kimbosoek/cosmos-images}

\noindent\textbf{Fingerprint} : Sokoto Coventry Fingerprint Dataset
\url{https://www.kaggle.com/datasets/ruizgara/socofing}

\noindent\textbf{Geometric shapes} : Geometric Shapes Mathematics
\url{https://www.kaggle.com/datasets/reevald/geometric-shapes-mathematics}

\noindent\textbf{Surface defect} : Surface Defect Detection Dataset
\url{https://www.kaggle.com/datasets/yidazhang07/bridge-cracks-image}

\noindent\textbf{Cotatenis sneakers} : Sneakers Images from Several Brands like Nike, Adidas, and Jordan.
\url{https://www.kaggle.com/datasets/ferraz/cotatenis-sneakers}

\noindent\textbf{Graphs} : About 16k of Clean Images of Graphs
\url{https://www.kaggle.com/datasets/sunedition/graphs-dataset}

\noindent\textbf{Handwritten digits and operators} : Handwritten Digits and Operators Dataset
\url{https://www.kaggle.com/datasets/sunedition/graphs-dataset}

\noindent\textbf{Insect village synthetic} : Dataset Containing Synthetic Images of Insects within Varying Backgrounds
\url{https://www.kaggle.com/datasets/vencerlanz09/insect-village-synthetic-dataset}

\noindent\textbf{Eye diseases} : Eye Diseases Retinal Images
\url{https://www.kaggle.com/datasets/gunavenkatdoddi/eye-diseases-classification}

\noindent\textbf{CAPTCHA} : Alphanumeric Colorful Images
\url{https://www.kaggle.com/datasets/parsasam/captcha-dataset}

\noindent\textbf{Kvasir} : Multi-class Image Dataset for Computer Aided Gastrointestinal Disease Detection
\url{https://www.kaggle.com/datasets/meetnagadia/kvasir-dataset}

\noindent\textbf{Meat quality} : Meat Quality Assessment Dataset
\url{https://www.kaggle.com/datasets/crowww/meat-quality-assessment-based-on-deep-learning}

\noindent\textbf{Chicken disease} : Machine Learning Dataset for Poultry Diseases Diagnostics
\url{https://www.kaggle.com/datasets/allandclive/chicken-disease-1}

\noindent\textbf{Facebook meme} : Facebook Hateful Meme Dataset
\url{https://www.kaggle.com/datasets/parthplc/facebook-hateful-meme-dataset}

\noindent\textbf{Blossom} : Hackathon Blossom (Flower Classification)
\url{https://www.kaggle.com/datasets/spaics/hackathon-blossom-flower-classification}

\noindent\textbf{Synthetic digits} : Synthetically Generated Images of English Digits Embedded on Random Backgrounds
\url{https://www.kaggle.com/datasets/prasunroy/synthetic-digits}

\noindent\textbf{AFFiNe} : Angling Freshwater Fish Netherlands
\url{https://www.kaggle.com/datasets/jorritvenema/affine}

\noindent\textbf{Fight} : Fight dataset
\url{https://www.kaggle.com/datasets/anbumalar1991/fight-dataset}

\noindent\textbf{Fish species} : Fish Species Image Data
\url{https://www.kaggle.com/datasets/sripaadsrinivasan/fish-species-image-data}

\noindent\textbf{Anime face} : 21551 Anime Face Images Sctaped from Web
\url{https://www.kaggle.com/datasets/soumikrakshit/anime-faces}

\noindent\textbf{ECG} : ECG Image Data
\url{https://www.kaggle.com/datasets/erhmrai/ecg-image-data}

\noindent\textbf{Tea disease} : Dieasese in Tea Leaves Image Classification
\url{https://www.kaggle.com/datasets/shashwatwork/identifying-disease-in-tea-leafs}

\noindent\textbf{Google scraped} : Google Scraped Image Dataset
\url{https://www.kaggle.com/datasets/duttadebadri/image-classification}

\noindent\textbf{Split garbage} : Split Garbage Dataset
\url{https://www.kaggle.com/datasets/andreasantoro/split-garbage-dataset}

\noindent\textbf{WebScreenshots} : Web Pages Classified by Their Screenshots
\url{https://www.kaggle.com/datasets/aydosphd/webscreenshots}

\noindent\textbf{ALL} : Acute Lymphoblastic Leukemia (ALL) Image Dataset
\url{https://www.kaggle.com/datasets/mehradaria/leukemia}

\noindent\textbf{Hand} : Hand Gesture Recognition Database
\url{https://www.kaggle.com/datasets/gti-upm/leapgestrecog}

\noindent\textbf{House price} : House Prices and Images
\url{https://www.kaggle.com/datasets/ted8080/house-prices-and-images-socal}

\noindent\textbf{Fast food} : Fastfood Image Classification
\url{https://www.kaggle.com/datasets/ganesh124/fastfood}

\noindent\textbf{Kannada} : Kannada Handwritten Characters
\url{https://www.kaggle.com/datasets/dhruvildave/kannada-characters}

\noindent\textbf{Logos} : Logos of BK, KFC, McDonald, Starbucks and Subway
\url{https://www.kaggle.com/datasets/kmkarakaya/logos-bk-kfc-mcdonald-starbucks-subway-none}

\noindent\textbf{Fruit and vegetable} : Fruits, Vegetables and Flowers for Image Classification and Object Detection
\url{https://www.kaggle.com/datasets/tobiek/green-finder}

\noindent\textbf{Mars} : Mars Surface and Curiosity Image Set
\url{https://www.kaggle.com/datasets/brsdincer/mars-surface-and-curiosity-image-set-nasa}

\noindent\textbf{YouTube} : YouTube Thumbnail Dataset
\url{https://www.kaggle.com/datasets/praneshmukhopadhyay/youtube-thumbnail-dataset}

\noindent\textbf{Book} : Book covers dataset
\url{https://www.kaggle.com/datasets/lukaanicin/book-covers-dataset}

\noindent\textbf{Belgium traffic} : Belgium Traffic Signs
\url{https://www.kaggle.com/datasets/shazaelmorsh/trafficsigns}

\noindent\textbf{Traffic light} : Traffic Light Detection Dataset
\url{https://www.kaggle.com/datasets/wjybuqi/traffic-light-detection-dataset}

\noindent\textbf{Montreal parking} : Montreal Parking Hours per Street
\url{https://www.kaggle.com/datasets/alincijov/montreal-parking-hours-per-streetsign}

\noindent\textbf{Wild plants} : Wild Plants Image Dataset
\url{https://www.kaggle.com/datasets/gverzea/edible-wild-plants}

\noindent\textbf{Pests} : Pests Identification
\url{https://www.kaggle.com/datasets/abhinandanroul/pest-normalized}

\noindent\textbf{Cattle} : Cattle Breeds Dataset
\url{https://www.kaggle.com/datasets/anandkumarsahu09/cattle-breeds-dataset}

\noindent\textbf{Fake video} : Fake Video Images Dataset
\url{https://www.kaggle.com/datasets/volodymyrgavrysh/fake-video-images-dataset}

\noindent\textbf{ChestX} : Chest Xrays Image Dataset
\url{https://www.kaggle.com/datasets/kostasdiamantaras/chest-xrays-bacterial-viral-pneumonia-normal}

\noindent\textbf{Brain CT} : Brain CT Images with Intracranial Hemorrhage Masks
\url{https://www.kaggle.com/datasets/vbookshelf/computed-tomography-ct-images}

\noindent\textbf{PANDA} : Prostate Cancer Grade Assessment (PANDA) Challenge
\url{https://www.kaggle.com/datasets/xhlulu/panda-resized-train-data-512x512}

\noindent\textbf{Concrete surface} : Concrete Surface Image Processed with Match Filter
\url{https://www.kaggle.com/datasets/ahsanulislam/concrete-surface-image-filtered-with-match-filter}

\noindent\textbf{Apple} : Apple Products Image Dataset
\url{https://www.kaggle.com/datasets/radvian/apple-products-image-dataset}

\noindent\textbf{Surgical} : Labeled Surgical Tools and Images
\url{https://www.kaggle.com/datasets/dilavado/labeled-surgical-tools}

\noindent\textbf{Wind turbines} : Airbus Wind Turbines Patches
\url{https://www.kaggle.com/datasets/airbusgeo/airbus-wind-turbines-patches}

\noindent\textbf{Apparel} : Apparel images dataset
\url{https://www.kaggle.com/datasets/trolukovich/apparel-images-dataset}

\noindent\textbf{Blood cell} :Blood Cell Images
\url{https://www.kaggle.com/datasets/paultimothymooney/blood-cells}

\noindent\textbf{Caltech101} : Caltech-101 Dataset Contains of 9,146 Images from 101 Object Categories
\url{https://www.kaggle.com/datasets/imbikramsaha/caltech-101}

\noindent\textbf{Diabetic} : Google Diabetic Rateinopathy
\url{https://www.kaggle.com/datasets/sohaibanwaar1203/diabetic-rateinopathy-full}

\noindent\textbf{Toxic plant} : Toxic Plant Image Dataset
\url{https://www.kaggle.com/datasets/hanselliott/toxic-plant-classification}

\noindent\textbf{League logo} : English Premier League Logo Detection
\url{https://www.kaggle.com/datasets/alexteboul/english-premier-league-logo-detection-20k-images}

\noindent\textbf{COVID19} : COVID-19 Radiography Database
\url{https://www.kaggle.com/datasets/tawsifurrahman/covid19-radiography-database}

\noindent\textbf{Lunar rock} : Lunar Rock Image Classification
\url{https://www.kaggle.com/datasets/pranshu29/lunar-rock}

\noindent\textbf{Noisy number} : Noisy, Single-Digit Captcha Images
\url{https://www.kaggle.com/datasets/kadenm/noisy-digitbased-captcha-images}

\noindent\textbf{QR} : Benign and Malicious QR Codes
\url{https://www.kaggle.com/datasets/samahsadiq/benign-and-malicious-qr-codes}

\noindent\textbf{Pistol detection} : Pistol Detection
\url{https://www.kaggle.com/datasets/vaibhavtalekar/pistol-classification}

\noindent\textbf{Moth} : Moths Image Dataset Classification
\url{https://www.kaggle.com/datasets/gpiosenka/moths-image-datasetclassification}

\noindent\textbf{Clock} : Dataset Containing 50K Generated Images of Analog Clocks
\url{https://www.kaggle.com/datasets/shivajbd/analog-clocks}

\noindent\textbf{Tomato} : Tomato Leaf Disease Image Classification
\url{https://www.kaggle.com/datasets/noulam/tomato}

\noindent\textbf{Lemon} : Lemon Quality Dataset
\url{https://www.kaggle.com/datasets/yusufemir/lemon-quality-dataset}

\noindent\textbf{IoT signal} : IoT Firmware Image Classification
\url{https://www.kaggle.com/datasets/datamunge/iot-firmware-image-classification}

\noindent\textbf{PLD} : Potato Disease Leaf Dataset(PLD)
\url{https://www.kaggle.com/datasets/rizwan123456789/potato-disease-leaf-datasetpld}

\noindent\textbf{Crocodile} : Crocodile|Alligator|Gharial Classification
\url{https://www.kaggle.com/datasets/rrrohit/crocodile-gharial-classification-fastai}

\end{document}